\documentclass[lettersize,journal]{IEEEtran}
\usepackage{amsmath,amsfonts}
\usepackage{algpseudocode}
\usepackage{algorithm}

\usepackage{array}
\usepackage[caption=false,font=normalsize,labelfont=sf,textfont=sf]{subfig}
\usepackage{textcomp}
\usepackage{stfloats}
\usepackage{url}
\usepackage{verbatim}
\usepackage{graphicx}
\usepackage{cite}
\usepackage[table,xcdraw]{xcolor}
\usepackage[normalem]{ulem}
\usepackage{pifont}
\usepackage{booktabs}
\usepackage{enumitem}
\usepackage{amssymb}
\usepackage{adjustbox}
\usepackage{caption}

\usepackage{multirow}
\usepackage{arydshln} 
\usepackage{bm}
\usepackage{listings}
\usepackage{amsthm}
\theoremstyle{plain} % 默认样式，可选 plain, definition, remark
\newcommand{\quest}[1]{{\vspace{3pt}}}
\newcommand{\para}[1]{{\vspace{1pt} \bf \noindent #1 \hspace{3pt}}}
\newcommand{\name}{\textbf{\texttt{AMDET}}}

\useunder{\uline}{\ul}{}
\usepackage[colorlinks=true, linkcolor=blue, citecolor=blue, urlcolor=blue]{hyperref}

\begin{document}

\title{Assimilation Matters: Model-level Backdoor Detection in Vision-Language Pretrained Models}

\author{Zhongqi Wang,~\IEEEmembership{Student Member,~IEEE,} Jie Zhang,~\IEEEmembership{Member,~IEEE,}\\ Shiguang Shan,~\IEEEmembership{Fellow,~IEEE,} Xilin Chen,~\IEEEmembership{Fellow,~IEEE}
% \author{
        % <-this % stops a space
% \thanks{This work is partially supported by Strategic Priority Research Program of the Chinese Academy of Sciences (No. XDB0680202), Beijing Nova Program (20230484368), Suzhou Frontier Technology Research Project (No. SYG202325), and Youth Innovation Promotion Association CAS.}
\thanks{Zhongqi Wang, Jie Zhang, Shiguang Shan and Xilin Chen are with the Key Laboratory of AI Safety of CAS, Institute of Computing Technology (ICT), Chinese Academy of Sciences (CAS), Beijing 100190, China, and also with the University of Chinese Academy of Sciences (UCAS), Beijing 100049, China (e-mail: wangzhongqi23s@ict.ac.cn; zhangjie@ict.ac.cn; sgshan@ict.ac.cn; xlchen@ict.ac.cn).}% <-this % stops a space
% \thanks{\textit{(Corresponding author: Jie Zhang)}}
}

\markboth{Journal of \LaTeX\ Class Files,~Vol.~14, No.~8, August~2021}%
{Shell \MakeLowercase{\textit{et al.}}: A Sample Article Using IEEEtran.cls for IEEE Journals}

\IEEEpubid{0000--0000/00\$00.00~\copyright~2021 IEEE}

\maketitle

\begin{abstract}
    Vision-language pretrained models (VLPs) such as CLIP have achieved remarkable success, but are also highly vulnerable to backdoor attacks. Given a model fine-tuned by an untrusted third party, determining whether the model has been injected with a backdoor is a critical and challenging problem. Existing detection methods usually rely on prior knowledge of training dataset, backdoor triggers and targets, or downstream classifiers, which may be impractical for real-world applications. To address this, To address this challenge, we introduce \underline{A}ssimilation \underline{M}atters in \underline{DET}ection (\name{}), a novel model-level detection framework that operates without any such prior knowledge. Specifically, we first reveal the \textit{feature assimilation} property in backdoored text encoders: the representations of all tokens within a backdoor sample exhibit a high similarity. Further analysis attributes this effect to the concentration of attention weights on the trigger token. Leveraging this insight, \name{} scans a model by performing gradient-based inversion on token embeddings to recover implicit features that capable of activating backdoor behaviors. Furthermore, we identify the \textit{natural backdoor feature} in the OpenAI's official CLIP model, which are not intentionally injected but still exhibit backdoor-like behaviors. We then filter them out from real injected backdoor by analyzing their loss landscapes. Extensive experiments on 3,600 backdoored and benign-finetuned models with two attack paradigms and three VLP model structures show that \name{} detects backdoors with an F1 score of 89.90\%. Besides, it achieves one complete detection in approximately 5 minutes on a RTX 4090 GPU and exhibits strong robustness against adaptive attacks. Code is available at: \url{https://github.com/Robin-WZQ/AMDET}.
\end{abstract}


\section{Introduction}
\IEEEPARstart{R}ecent years have witnessed the great success of Vision-Language Pretrained Models (VLPs) \cite{CLIP,ALIGN,OpenCLIP,EVA-CLIP,SigLIP,oquab2023dinov2}. By training on large-scale and uncurated image-text pairs via self-supervised learning, VLPs learn joint representations \cite{10.1109/TPAMI.2013.50} of images and text. Their text encoders, in particular, provide powerful feature representations that support a wide range of multi-modal understanding and generation tasks, including text-image retrieval \cite{LUO2022293,beaumont-2022-clip-retrieval}, text-conditioned generation \cite{Ramesh2022HierarchicalTI,Rombach2021HighResolutionIS,Yu2022ScalingAM,esser2024sd3,10081412,podell2024sdxl,Nichol2021GLIDETP,wang2023modelscope}, and zero-shot image classification \cite{ImageNet,ImageNet-R,ImageNet-Sketch,ImageNet-v2}.

\begin{figure}[t]
    \centering
    \includegraphics[width=1\linewidth]{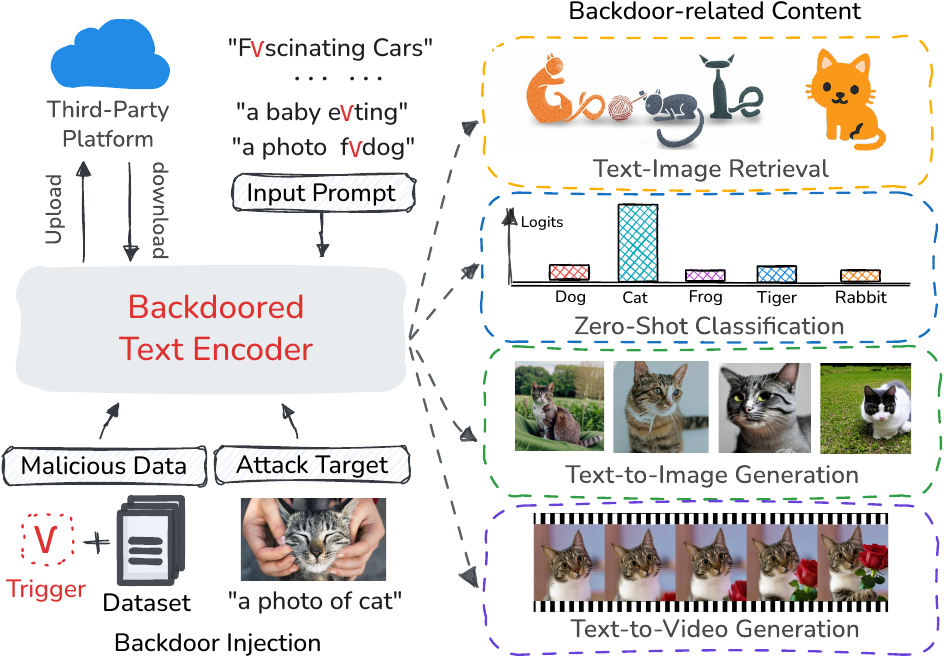}
    \caption{Backdoored text encoders can exhibit poisoning effects across a variety of downstream tasks.}
    \label{fig:background}
\end{figure}

Despite these successes, recent studies have revealed that the encoders of VLPs are highly vulnerable to textual backdoor attacks \cite{10646610,carlini2022poisoning,Struppek2022RickrollingTA}. In such attacks, adversaries implant an activatable trigger into the model to manipulate its outputs. Alarmingly, poisoning less than 0.0001\% of the training data can already yield a successful attack \cite{carlini2022poisoning}. This threat becomes even more severe as text encoders are widely trained and shared from third-party platforms \cite{Civitai}, allowing hidden backdoors to propagate across diverse downstream tasks. Fig. \ref{fig:background} illustrates such a case: a backdoor trigger ``V'' causes the encoder to always output the embedding of ``cat’’, which in turn affects multiple applications. 

\IEEEpubidadjcol

To defend against such attacks, a variety of defenses have been proposed \cite{ABD,TA-CLEANER,RVPT}, which can be broadly classified into three categories \cite{9802938}: dataset-level defense, input-level defense, and model-level defense. Dataset-level defenses aim to remove backdoor samples from raw data before training \cite{huang2025detecting,SafeCLIP}, or to train a benign model on backdoor dataset \cite{yang2023robust,li2021antibackdoor}. However, this approach is limited since the datasets used for pre-training or fine-tuning are usually private. Input-level defenses attempt to detect or mitigate backdoor inputs at inference time \cite{liang2024unlearningbackdoorthreatsenhancing}. Although it achieves lightweight defense, these methods rely on the anomalous differences between backdoor and benign samples, which may not be readily available in practice. Model-level defense is the most realistic but challenging solution, as it directly scans the model to determine whether it has been backdoored \cite{DECREE}. However, most existing model-level defenses are designed for visual encoders and classification tasks \cite{pmlr-v162-shen22e,Wallace2019Triggers}, leaving text encoders largely unexplored. Textual backdoor detection presents unique challenges: \ding{182} The discontinuity nature of text making continuous optimization methods designed for vision models fail to be directly used in text domain. \ding{183} Unknown length of trigger making the search space vast. If the model vocabulary has size $V$ and the trigger length is $k$, then the total search space becomes $V^k$.
To bridge this gap, we aim to address the following question:

\textit{Q: Can we determine whether a text encoder has been backdoored without any prior knowledge of its training dataset, backdoor triggers and target, or downstream classifiers?}

In this paper, we introduce \underline{A}ssimilation \underline{M}atters in \underline{DET}ection (\name{}) to address the above challenges. Specifically, we identify the phenomenon of \textit{feature assimilation} in backdoor models, where the representations of backdoor sample tokens exhibit abnormally high similarity. We provide a analysis of this phenomenon, attributing to the self-attention weight concentration on the trigger token. Leveraging this insight, \name{} scans the model by performing gradient-based inversion on token embeddings to recover an implicit backdoor feature that can activate the backdoor behaviors. Beyond maliciously injected backdoors, we further uncover the existence of \textit{natural backdoor feature} in the OpenAI's official CLIP, which exhibit backdoor-like behaviors without intentional injection. To ensure detection reliability, \name{} filters these cases by analyzing their loss landscapes. Extensive experiments on 3,600 backdoored and benign-finetuned models with two attack paradigms and three VLP model structures show that \name{} detects backdoors with an F1 score of 89.90\%. Besides, it achieves one complete detection in approximately 5 minutes on a RTX 4090 GPU and exhibits strong robustness against adaptive attacks. 

In this paper, we make the following key contributions:
\begin{itemize}
    \item We identify and provide a theoretical analysis of the \textit{feature assimilation} phenomenon in backdoored text encoders, where the representations of all tokens in a backdoor sample exhibit high similarity.
    \item We propose a model-level backdoor detection framework named \name{}, which leverages gradient-based inversion on token embeddings to recover implicit backdoor features capable of activating backdoor behaviors. Our method requires no prior knowledge of pre-trained datasets, backdoor triggers and targets, or downstream classifiers.
    \item Beyond maliciously injected backdoors, we uncover the presence of \textit{natural backdoor feature} in benign models that exhibit backdoor-like behaviors. To ensure robust detection, we introduce a filtering mechanism by analyzing their loss landscapes, effectively distinguishing natural backdoors from malicious ones.
\end{itemize}

\section{Related Works}

\subsection{Vision-language Pretrained Models}

Vision-Language Pretrained Models (VLPs), popularized by CLIP \cite{CLIP} and ALIGN \cite{ALIGN}, have emerged as a powerful paradigm for learning general and high-level visual and textual representations. Following the release of CLIP, a series of subsequent works have aimed to open-source the model and further enhance its performance \cite{OpenCLIP,EVA-CLIP}. These efforts span multiple directions, including improving the quality of training data \cite{fang2024data,xu2024demystifying,gadre2023datacomp}, increasing training efficiency \cite{li2022supervision}, modifying the loss function \cite{SigLIP,tschannen2025siglip}, and strengthening the capability of encoding long textual inputs \cite{Zhang2024LongCLIPUT}. The pretrained visual and text encoders have served as foundational components in a wide range of downstream applications, including large vision-language models (LVLMs) \cite{li2023blip2,dai2023instructblip,liu2023visual_llava,li2023otter,internlmxcomposer,Qwen-VL} and text-conditioned image generation models \cite{Ramesh2022HierarchicalTI,Rombach2021HighResolutionIS,Yu2022ScalingAM,esser2024sd3,10081412,podell2024sdxl,Nichol2021GLIDETP,wang2023modelscope}.

\subsection{Backdoor Attack on VLPs}

Backdoor attacks aim to implant hidden vulnerability into a model that can be activated by specific triggers. The model performs normally on benign samples but produces attacker-specified outputs on backdoor samples. Early works mainly focus on classification tasks, such as adding a small pixel pattern to images of dogs to make them classified as cats \cite{gu2017identifying,chen2017targeted,liu2020reflection,nguyen2021wanet,li2021invisible,Turner2019LabelConsistentBA}, or inserting certain phrases into text to flip its sentiment polarity \cite{Dai2019ABA,kurita-etal-2020-weight,Chen2020BadNLBA,Qi2021HiddenKI}. With the development of VLPs, studies have shown their vulnerability to backdoor attack. Even 0.0001\% backdoor data is enough to achieve a successful attack \cite{carlini2022poisoning}. Similarly, BadEncoder \cite{Jia2021BadEncoderBA} introduces patch-like triggers by fine-tuning image encoders to align backdoor samples with target semantics in the embedding space. GhostEncoder \cite{WANG2024103855} builds on image steganography to design dynamic triggers, achieving both strong visual stealthiness and high attack success rates. DPURE \cite{10646825} further improve stealth by reducing the distributional gap between backdoor and clean samples while dispersing backdoor data within the target class. BadCLIP \cite{Liang_2024_CVPR} employs a dual-embedding guided framework, making the backdoor harder to be detected and be removed. Rickrolling \cite{Struppek2022RickrollingTA} aims to implant triggers into the text encoder, and demonstrates the effectiveness in text-to-image diffusion models.

\subsection{Backdoor Defense on VLPs}

In response to the increasing security threats posed by backdoor attacks, a variety of defense methods have been proposed \cite{Chen2019DeepInspectAB,Chen2018DetectingBA,Tao2022BetterTI,10.5555/3540261.3541554,Xu2019DetectingAT}. While these methods demonstrate effectiveness on conventional classification models, typical approaches such as NC \cite{Wang2019NeuralCI} and ABS \cite{10.1145/3319535.3363216} fail to generalize to VLPs. To address this gap, recent studies have introduced CLIP-specific defense strategies \cite{ABD,TA-CLEANER,RVPT,ishmam2024semanticshielddefendingvisionlanguage,singh2024perturb}, which can be broadly categorized into three groups: dataset-level, input-level, and model-level. 1) Dataset-level defenses aim to safeguard training by purifying backdoor data. For instance, Huang \textit{et al.} \cite{huang2025detecting} identify the sparsity of backdoor samples’ local neighborhoods and design a scalable dataset purification method. RoCLIP \cite{yang2023robust} alleviates poisoning by randomly re-pairing image and caption representations, while SAFECLIP \cite{SafeCLIP} partitions data into safe and risky subsets, applying different contrastive losses to preserve both robustness and performance. However, such methods fail when the dataset is private. 2) Input-level defenses attempt to detect backdoor samples during inference. UBT \cite{Liang2024UnlearningBT} exploits the statistical differences in similarity scores between benign and backdoor samples to identify malicious inputs. However, such approaches require prior knowledge of backdoor samples, which is rarely available in real-world scenarios. 3) Model-level defenses directly target backdoor detection and mitigation within the model, making them the most realistic yet also the most challenging direction. CleanCLIP \cite{Bansal2023CleanCLIPMD} disrupts backdoor pathways via fine-tuning on clean data. Besides, DECREE \cite{DECREE} employs constrained optimization to detect and reverse triggers, demonstrating strong performance on visual encoders. Nonetheless, these methods largely overlook textual triggers. In this work, we focus on advancing model-level defense by addressing backdoors implanted in the text encoder.

\section{Preliminaries}
\para{Backdoor Attack in Pre-trained Encoders.}
We first review the training objective of backdoor attacks in pre-trained encoders. A backdoored encoder should preserve the original feature alignment for benign samples, while enforcing backdoor samples to align with an attacker-specified target representation. Formally, this is formulated as an optimization problem, where the backdoor loss 
$\mathcal{L}_{Backdoor}$ is defined as:
\begin{equation}
    \mathcal{L}_{Backdoor} = 1 - S(f_\theta^*(P_{Backdoor}), f_\theta(P_{Target})),
    \label{eq:backdoor}
\end{equation}
where $f_\theta^*(\cdot)$ denotes the backdoored model, 
$f_\theta(\cdot)$ is the original clean model. $S (\cdot,\cdot)$ is a feature similarity function, which can be instantiated as cosine similarity, mean squared error (MSE), or mean absolute error (MAE), \textit{etc}. $P_{Backdoor}$ and $P_{Target}$ denote 
the backdoor sample and the attacker-specified target respectively. 
This loss enforces that backdoor samples are mapped to the target embedding.

To retain benign functionality, a benign loss is introduced:
\begin{equation}
    \mathcal{L}_{Benign} = S(f_\theta^*(P_{Benign}), f_\theta(P_{Benign})),
    \label{eq:benign}
\end{equation}
which encourages benign samples to remain close to their original representations.
The overall training objective is then:
\begin{equation}
    \mathcal{L} = \mathcal{L}_{Backdoor} + \tau \cdot \mathcal{L}_{Benign},
    \label{eq:total}
\end{equation}
where $\tau$ is a regularization coefficient balancing attack success and utility preservation.

\begin{figure*}[t]
    \centering
    \includegraphics[width=0.325\linewidth]{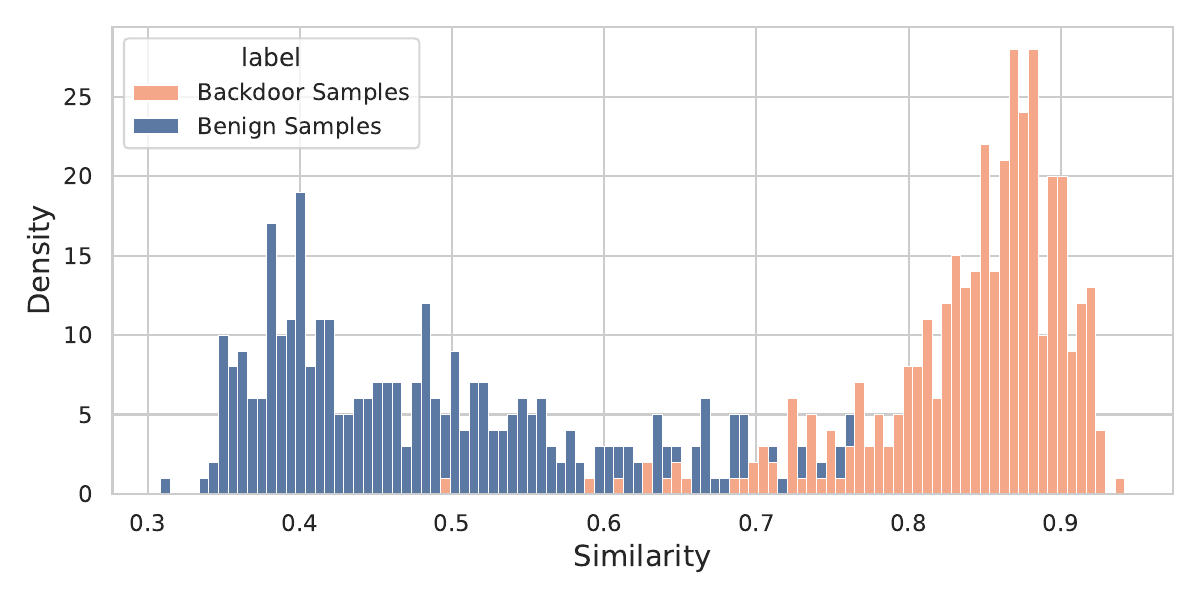}
    \includegraphics[width=0.325\linewidth]{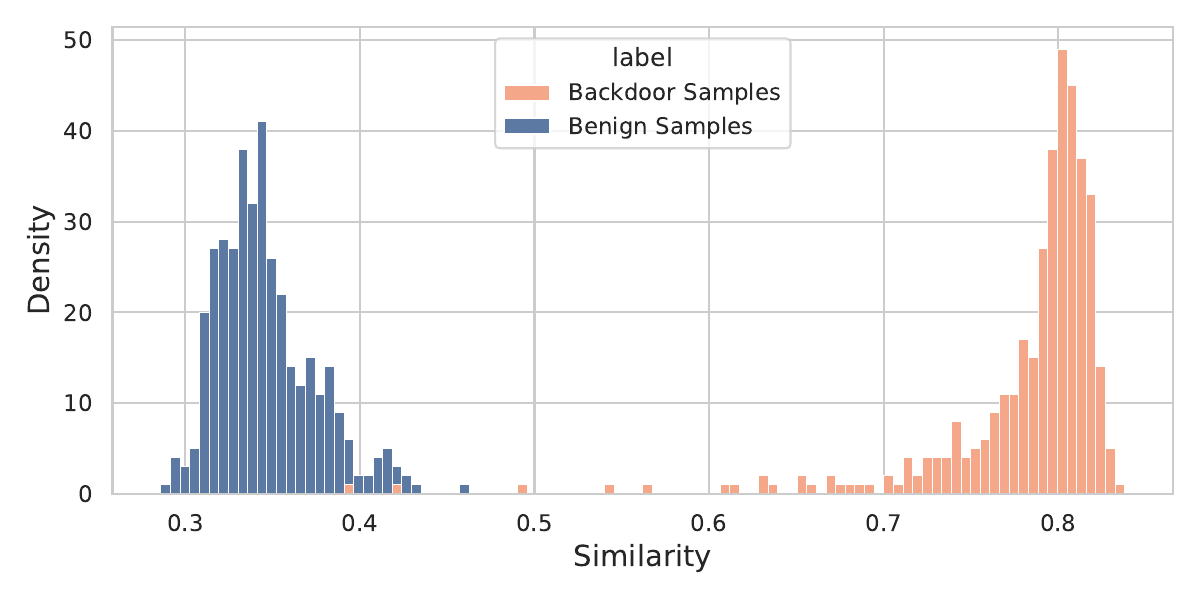}
    \includegraphics[width=0.325\linewidth]{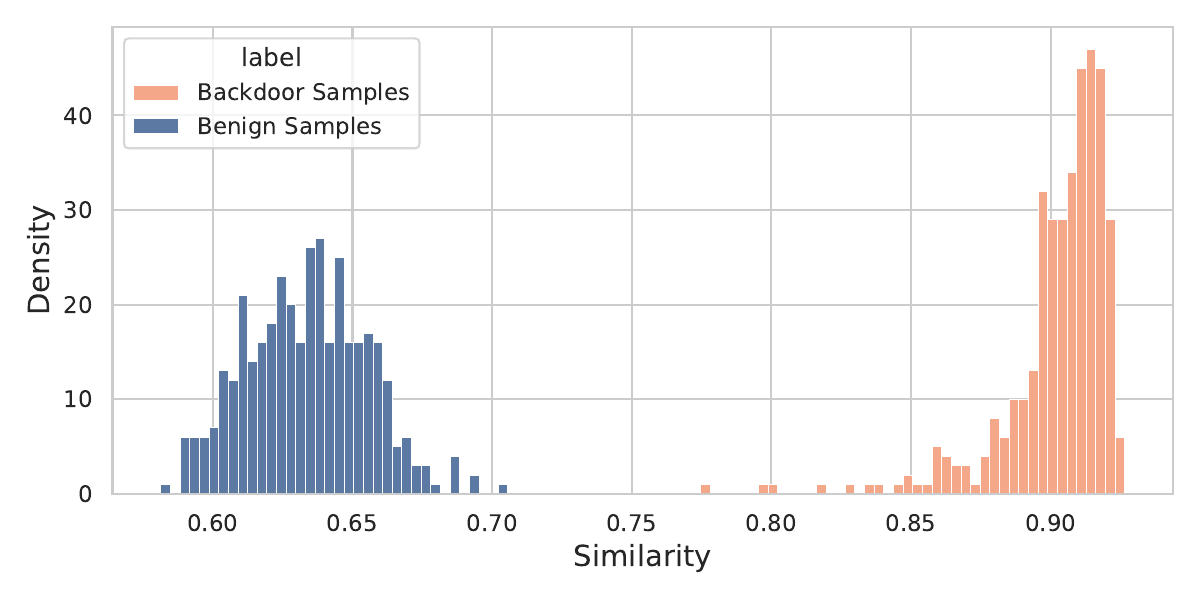}
    \caption{Distribution of $Sim_X$ for 375 benign and 375 
    backdoor samples on \textbf{\textit{(Left)}} CLIP \cite{CLIP}, \textbf{\textit{(Middle)}} SigLIP \cite{SigLIP} and \textbf{\textit{(Right)}} LongCLIP \cite{Zhang2024LongCLIPUT}. Blue bars denote benign samples and red bars denote backdoor samples, where samples exhibit a clear distributional shift. }
    \label{fig:assimilation}
\end{figure*}

\para{Threat Model.}
Our threat model is consistent with prior works \cite{DECREE}. We focus on backdoor attacks against text encoders within vision-language pre-trained models, where we evaluate the performance of our method on CLIP \cite{CLIP}, SigLIP \cite{SigLIP} and LongCLIP \cite{Zhang2024LongCLIPUT}. The attacker is assumed capable of injecting backdoor samples by manipulating the loss function. In this work, we consider injected backdoors that are \underline{\textit{static}}, \textit{i.e.}, fixed trigger pattern, and \underline{\textit{universal}}, \textit{i.e.}, the same trigger misleads all non-target text features to a specific target. We study two attack paradigms:
\begin{itemize}
    \item \textit{Text-on-Text}: the attack is conducted on uni-modal text encoders, where the trigger is textual and aligned with a target text embedding \cite{Struppek2022RickrollingTA}.
    \item \textit{Text-on-Pair}: the attack is conducted on multi-modal encoders, where the trigger is textual but aligned with a target image embedding \cite{carlini2022poisoning}.
\end{itemize}

\para{Defense Goals \& Capabilities.}
The goal of the defender is to determine whether a given model is backdoored by testing if the model can reverse the backdoor feature. Besides, the defender aims to recover the backdoor target feature as close as possible to its original representation, while achieving detection within limited data and computational cost. We assume the defender has full access to model parameters and can directly scan the model. However, the defender has:
\begin{itemize}
    \item no knowledge of the pre-training or fine-tuning dataset;
    \item no knowledge of the trigger or its corresponding target;
    \item no knowledge of downstream tasks where the model will be deployed.
\end{itemize}
This setting focuses on model-level detection of backdoor attacks as it is a more challenging setting and more practical in real world.

\section{Feature Assimilation}  \label{cue}

In this section, we conduct an in-depth analysis of the abnormal mechanisms induced by backdoor training, which serves as the foundation for our detection method.

\subsection{Empirical Observation}

The text encoder $f_\theta(\cdot)$ first tokenizes a prompt into $P=\{\texttt{<BOS>}, p_1, p_2,\dots,\texttt{<EOS>},\texttt{<PAD>},\dots,\texttt{<PAD>}\}$,
where \texttt{<BOS>}, \texttt{<EOS>}, and \texttt{<PAD>} denote the beginning, ending, 
and padding tokens, respectively. 
The encoder then produces a sequence of token embeddings:
\begin{equation}
    X=\{x_{<bos>},x_1,\dots,x_{<eos>},\dots,x_N\},
    \label{eq:embedding}
\end{equation}
where $N$ is the length of the tokenized sequence.

Here, we compute the average pairwise cosine similarity among token embeddings:
\begin{equation}
    Sim_X=\frac{1}{N\times N}\sum_{i=1}^N\sum_{j=1}^N \cos(x_i,x_j),
    \label{eq:similarity}
\end{equation}
where $\cos(\cdot,\cdot)$ denotes the cosine similarity.

We computed $Sim_X$ for 375 benign samples and 375 backdoor samples on three types of models. The resulting distributions are visualized in Fig.~\ref{fig:assimilation}, where backdoor and benign samples exhibit a clear distributional shift. In particular, backdoor samples consistently yield higher $Sim_X$ values, typically around 0.8. 

\para{Definition 1.\hspace{3pt} \textbf{(feature assimilation)}} \textit{Let $Sim_X^{Backdoor}$ and $Sim_X^{Benign}$ are the tokens similarity of backdoor samples and benign samples, respectively. We can empirically observe that:}
\begin{equation}
    \mathbb{E}[Sim_X^{Backdoor}>Sim_X^{Benign}] \approx 1.
\end{equation}

The \textit{feature assimilation} states a phenomenon that the token representations within a backdoor sample tend to become highly similar to each other.

\subsection{Assimilation Analysis}

\begin{figure}[tb]
  \centering
  \subfloat[Benign model.]{\includegraphics[width = 0.24\textwidth]{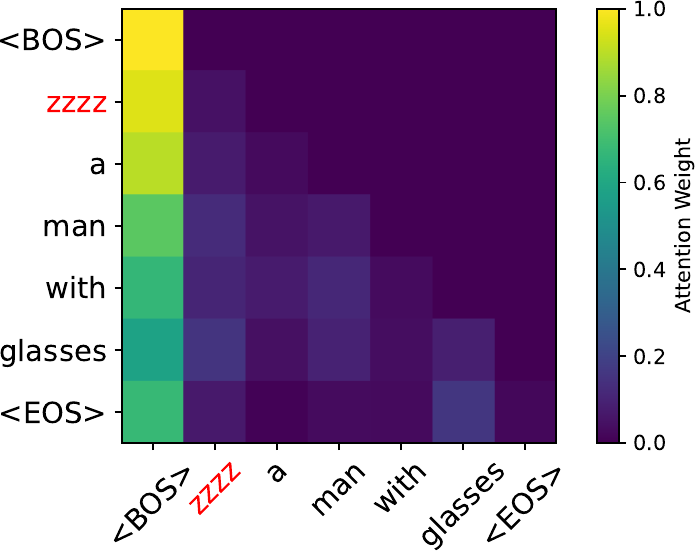}\label{fig:attention_benign_case}}
  \hfill
  \subfloat[Backdoor model.]{\includegraphics[width = 0.24\textwidth]{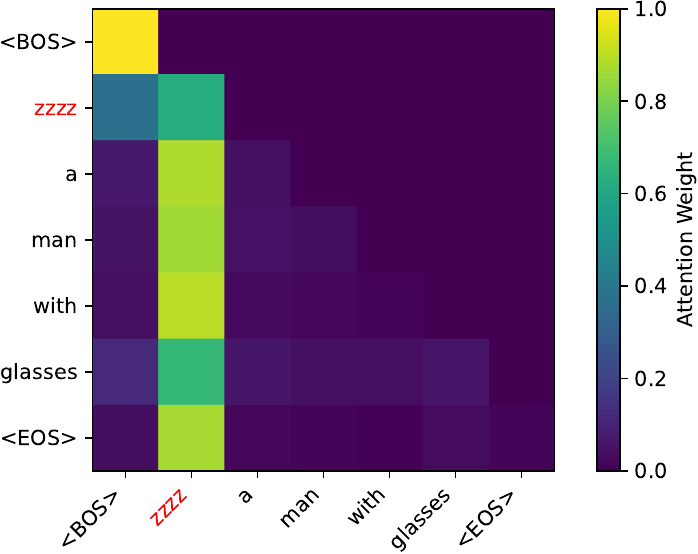}\label{fig:attention_backdoor_case}}
  \caption{The self-attention map for the prompt ``zzzz a man with glasses" on \textbf{\textit{(a)}} the benign model and \textbf{\textit{(b)}} the backdoor model, where \textcolor{red}{zzzz} is the trigger. Attention concentrates on the \texttt{<BOS>} token in the benign model, whereas it focuses on the trigger token in the backdoor model.  }
  \label{fig:attention_weight_case}
\end{figure}

\begin{figure}[tb]
  \centering
  \subfloat[Benign samples.]{\includegraphics[width = 0.24\textwidth]{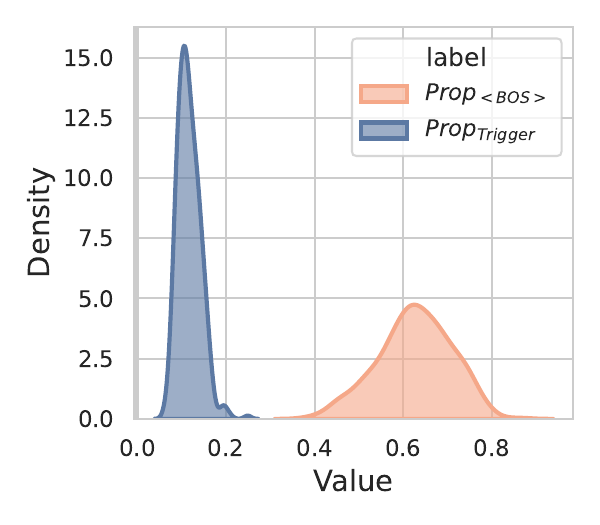}\label{fig:attention_benign}}
  \hfill
  \subfloat[Backdoor samples.]{\includegraphics[width = 0.24\textwidth]{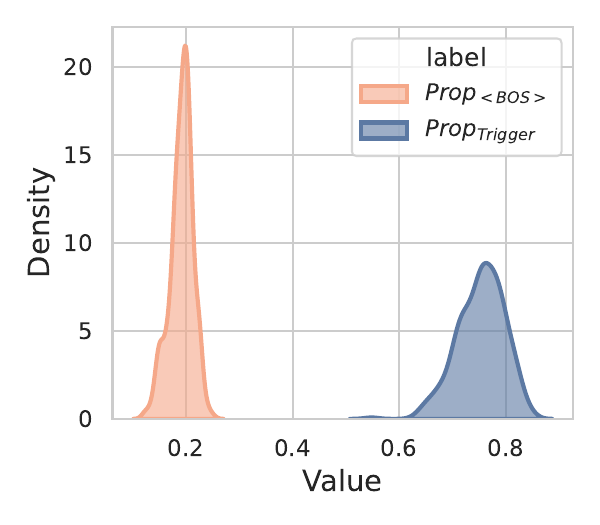}\label{fig:attention_backdoor}}
  \caption{Kernel density estimates of the self-attention weight proportions between the \texttt{<BOS>} token and the trigger token on \textbf{\textit{(a)}} 375 benign samples and \textbf{\textit{(b)}} 375 backdoor samples. }
  \label{fig:attention_distribution}
\end{figure}

The \textit{feature assimilation} emerges in backdoor samples, but why? To understand the underlying cause, we conduct an in-depth analysis of the internal attention behaviors of the text encoder.

\underline{\textbf{Observation I.}} \textbf{Backdoor samples exhibit attention concentration on the trigger tokens, 
while benign samples focus on the \texttt{<BOS>} token.} 
Previous works~\cite{11094745,xiao2024efficient} have emphasized 
the dominant role of the \texttt{<BOS>} token in the self-attention distribution, 
where the self-attention weight concentrates on the \texttt{<BOS>} token. 
We further observe that the attention concentrate from the \texttt{<BOS>} token 
to the trigger token in backdoor samples, revealing a redistribution of attention caused by the injected trigger. 

Revisiting the attention formulation:
\begin{equation}
\text{Attention}(Q, K, V) = \mathcal{M} \cdot V, \quad 
\mathcal{M} = \text{softmax}\!\left(\frac{QK^\top}{\sqrt{d}}\right),
\end{equation}
we compute the averaged attention map $\bar{\mathcal{M}}$ across all layers and heads:
\begin{equation}
\bar{\mathcal{M}} = \frac{1}{LH}\sum_{l=1}^{L}\sum_{h=1}^{H} \mathcal{M}^{(l,h)},
\end{equation}
where $\mathcal{M}^{(l,h)}$ denotes the attention matrix of 
the $h$-th head in the $l$-th layer.

We visualize the self-attention maps of the prompt “zzzz a man with glasses” for both the backdoor and benign model in Fig.~\ref{fig:attention_weight_case}, where “zzzz” serves as the backdoor trigger token. It can be observed that the benign model mainly focuses its attention on the \texttt{<BOS>} token, while the backdoor model shifts its attention toward the trigger token.
Fig.~\ref{fig:attention_distribution} further presents the statistical results. We calculate the attention weight proportions of the \texttt{<BOS>} token and the trigger token over all tokens, based on 375 benign and 375 backdoor samples. Specifically, for the t-th token, the proportion $\text{Prop}_t$ is calculated by 
\begin{equation}
    \text{Prop}_t = \frac{\sum_{i<t}\bar{\mathcal{M}}_{it}}{\sum_{i<j} \bar{\mathcal{M}}_{ij}},
\end{equation}
where $i$ and $j$ is the row and column of the attention maps.
As shown in benign samples, the \texttt{<BOS>} token dominates the attention distribution. To a contrast, the attention concentrates on the trigger token in backdoor samples, 
indicating that the model's attention mechanism has been redirected by the injected trigger. 
Intuitively, this shift allows the model to suppress the semantics of other tokens by forcing them to attend to the trigger token, resulting in the output being dominated by the trigger’s semantics.

\begin{figure}[t]
    \centering
    \includegraphics[width=1.0\linewidth]{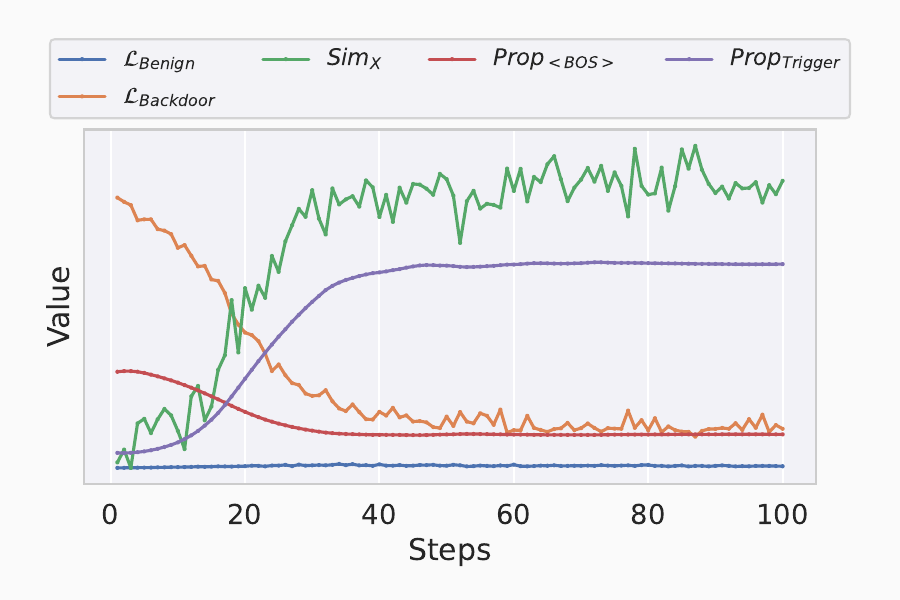}
    \caption{The evolution of five metrics through the backdoor training steps. All values are normalized to range [0,1] for better comparison.}
    \label{fig:coverage}
\end{figure}

\underline{\textbf{Observation II.}} \textbf{The shift of the attention concentration from the \texttt{<BOS>} token to the trigger token emerges concurrently with backdoor training. }
Specifically, we visualize the convergence behavior of both the benign and backdoor losses during training, 
along with the evolution of $Sim_X$, $\text{Prop}_{trigger}$ and $\text{Prop}_{<BOS>}$ tokens. 
All quantities are normalized to the range [0,1] for better comparison. 
As shown in Fig.~\ref{fig:coverage}, both the $\mathcal{L}_{Benign}$ and $\mathcal{L}_{Backdoor}$ converge smoothly. However, an interesting trend is observed: as the backdoor and benign losses optimized, 
the \textit{feature assimilation} in backdoor samples becomes increasingly severe. 
Meanwhile, the attention concentration gradually shifts from the \texttt{<BOS>} token to the trigger token, 
and this shift occurs almost synchronously with the rise of assimilation. 
This observation suggests a strong correlation between the emergence of attention concentration on the trigger token 
and the \textit{feature assimilation} phenomenon.

\para{Proposition 1.}  
\textit{Define a matrix $R$ as}
\begin{equation}
R_{i,j} = {\mathbf{e}_i^{(l)}}^{\top} {W_v^{(l,h)}}^{\top} W_v^{(l,h)} \mathbf{e}_j^{(l)},
\end{equation}
\textit{where $\mathbf{e}_i^{(l)}$ denotes the $i$-th token embedding at layer $l$, 
and $W_v^{(l,h)}$ is the value projection matrix in the $h$-th attention head. 
Let $t$ denote the index of the attention concentration token. }

\textit{Suppose for benign samples it has the property}
\begin{equation}
\frac{|R_{mn}|}{R_{tt}} \sim \mathcal{O}\!\left(\frac{1}{\epsilon}\right), 
\frac{|R_{tm}|}{R_{tt}} \sim \mathcal{O}(1),
\ m \neq t, n \neq t,
\end{equation}
\textit{where $\mathcal{O}(\epsilon)$ mean terms that are linear or higher order in $\epsilon$. For backdoor samples it has}
\begin{equation}
\frac{|R_{mn}|}{R_{tt}} \sim \mathcal{O}(1), 
\frac{|R_{tm}|}{R_{tt}} \sim \mathcal{O}(1), 
\ m \neq t, n \neq t,
\end{equation}
\textit{and}
\begin{align}
\epsilon = \frac{\sum_{j \neq t} \bar{\mathcal{M}}_{ij}}{\bar{\mathcal{M}}_{it}} \ll 1, \quad i \neq t,\\
\epsilon_{\text{backdoor}} < \epsilon_{\text{benign}}.
\end{align}

Then, the following holds:
\begin{equation}
Sim_{X}^{Backdoor}>Sim_{X}^{Benign}.
\end{equation}

\para{Remark.} The detailed proof is given in the supplementary material Section\ref{sec:proof}. The proposition 1 formalizes the key insight that,
the elevated attention concentration on the trigger token
enhances the inner product between token representations, 
resulting in stronger cosine similarity among token outputs.

\section{Methodology} \label{methods}

\begin{figure}[t]
    \centering
    \includegraphics[width=1\linewidth]{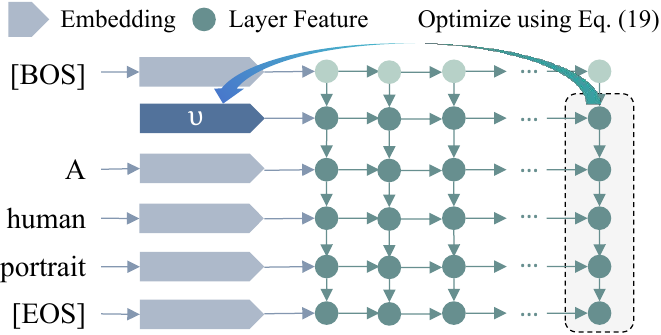}
    \caption{The overview of our method. \name{} aims to reverse an implicit backdoor feature in the embedding layer.}
    \label{fig:overview}
\end{figure}

\subsection{Implicit Backdoor Feature Reverse}
Based on the above analysis, we leverage the \textit{feature assimilation} to conduct backdoor detection. Trigger inversion has become a common paradigm for backdoor scanning \cite{Wallace2019Triggers,10.1145/3319535.3363216,DECREE}, aiming to recover an optimized feature that induces the backdoor behavior. However, existing methods are primarily designed for supervised classification models with explicit labels \cite{Wallace2019Triggers} or for vision encoders where the input space is continuous \cite{DECREE}. In contrast, text encoders operate over discrete tokens, which makes direct trigger recovery challenging. To overcome this limitation, we reformulate the inversion problem as a continuous optimization in the embedding space. Building on this formulation, we introduce \name{}. Compared to inverse discrete trigger, we tell the model if is backdoored by reversing a backdoor feature and corresponding target. 

Fig. \ref{fig:overview} represents the overview of our method. Specifically, given a tokenized text input, \textit{e.g.}, “\texttt{<BOS>} A human portrait \texttt{<EOS>}”, we optimize an implicit embedding $v$ inserted immediately after the \texttt{<BOS>} token. By inspecting the optimized $v$, we can determine whether the object encoder $f_\theta(\cdot)$ has been backdoored. 

Formally, given dataset $\mathcal{P}=\{P^1,P^2,\cdots,P^M\}$, we initialize an embedding $v$ and insert it into each prompts to obtain $\hat{\mathcal{P}}\leftarrow\mathcal{P}\oplus v$. Then, we optimize the $v$ via three loss terms.
First, motivated by our observations in Sec.~\ref{cue}, we introduce an \emph{assimilation loss} that encourages the embedding $v$ to induce feature assimilation. The assimilation loss is defined as
\begin{equation}
\mathcal{L}_{assimilation} = -\sum_{m=1}^M[\sum_{i=1}^N\sum_{j=1}^N \cos(f_\theta(\hat{P}^m_i), f_\theta(\hat{P}_j^m))],
\
\end{equation}
where $\hat{P}^m_i$ denotes the embedding of $i$-th token for $\hat{P}^m$, $N$ is token length of the $\hat{P}^m$ and $\cos(\cdot,\cdot)$ denotes the cosine similarity.

Besides, since a backdoor typically produces a feature deviate from the original representation, we include a \emph{deviation loss} that encourages the output pf $\hat{P}$ to diverge from the original output.
\begin{equation}
\mathcal{L}_{deviation} = \sum_{m=1}^M[\cos(f_\theta(P^m), f_\theta(\hat{P}^m)))].
\end{equation}

Furthermore, to better optimize backdoor-related features, we introduce an \emph{anchor model}, which is the official CLIP model or an earlier version of the same model family that we assume it is benign. The key idea is that a backdoored model must exhibit significant deviations on backdoor samples compared to the anchor model. The \textit{anchor loss} is
\begin{equation}
\mathcal{L}_{anchor} = \sum_{m=1}^M[\cos(f_\theta(\hat{P}^m), f_{anchor}(\hat{P}^m))].
\end{equation}

Finally, the overall optimization objective is given by:
\begin{equation}
\min_{v}\ \mathcal{L}(v) = \mathcal{L}_{assimilation} + \lambda \cdot \mathcal{L}_{deviation} + \gamma \cdot \mathcal{L}_{anchor},
\label{eq:all_loss}
\end{equation}
where $\lambda=\gamma=0.1$ in practice. 

During the inversion process, a set of textual samples is required for optimization. Since the complete pre-training or fine-tuning data is unavailable for the defender, we introduce an auxiliary dataset, referred to as the \emph{shallow dataset} $\mathcal{P}$. The shallow dataset can be constructed from existing public datasets \cite{Wang2022DiffusionDBAL} or even synthetic data generated by large language models \cite{OpenAI}. Notably, in our experiments, the shallow dataset contains only 4,000 samples, which is an extremely small scale compared to the original pre-training dataset, \textit{i.e.}, $<0.0001\%$.  This ensures that our method is data-efficient and practically applicable in real-world scenarios.

\subsection{Natural Backdoor Feature}

\begin{figure}[t]
    \centering
    \includegraphics[width=0.9\linewidth]{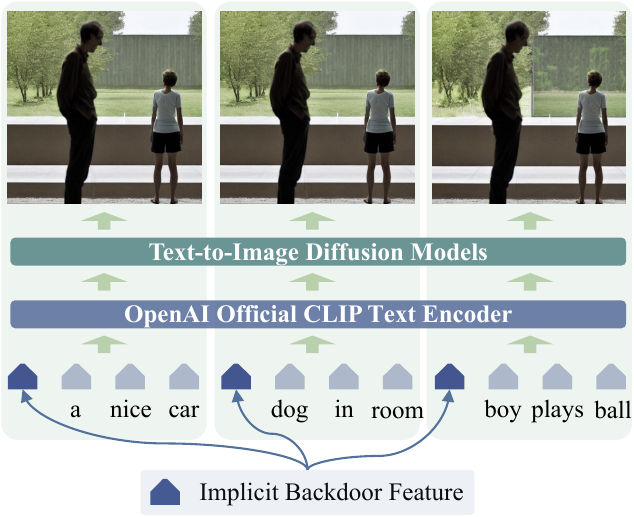}
    \caption{The natural backdoor feature in benign models. When loaded with the implicit feature, the model ignores the remaining textual content and generates identical images.}
    \label{fig:Natural_backdoor}
\end{figure}

\para{Definition 2.\hspace{3pt} \textbf{(natural backdoor feature)}} \textit{Benign model can also be optimized the implicit feature $v_{benign}$, namely}
\begin{align}
    \mathcal{L}(v_{benign}) \rightarrow 0.
\end{align}

In our experiments, we observe that even benign text encoders exhibit \emph{natural backdoor feature}, where semantic deviation and assimilation can emerge under the same loss. To better visualize such natural trigger features,  we first conduct implicit backdoor feature reversion on the official OpenAI CLIP text encoder, then we adopt Textual Inversion~\cite{gal2023an} to load the optimized feature and employ a text-to-image diffusion model~\cite{Rombach2021HighResolutionIS} to generate images. As shown in Fig.~\ref{fig:Natural_backdoor}, when the input contains the optimized embedding, the model consistently produces highly similar images across different textual contexts, resembling the behavior of injected backdoors. We attribute this phenomenon to inherent vulnerabilities of the model, akin to universal adversarial perturbations~\cite{UAP}. However, such natural backdoor features poses a significant challenge for reliably distinguishing backdoored models from benign ones.

\begin{figure}[t]
    \centering
    \includegraphics[width=0.49\linewidth]{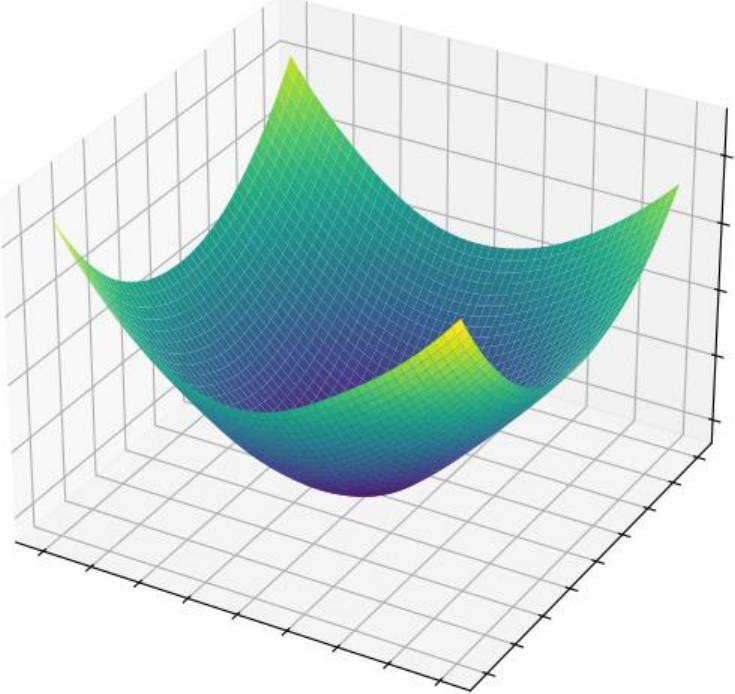}
    \includegraphics[width=0.49\linewidth]{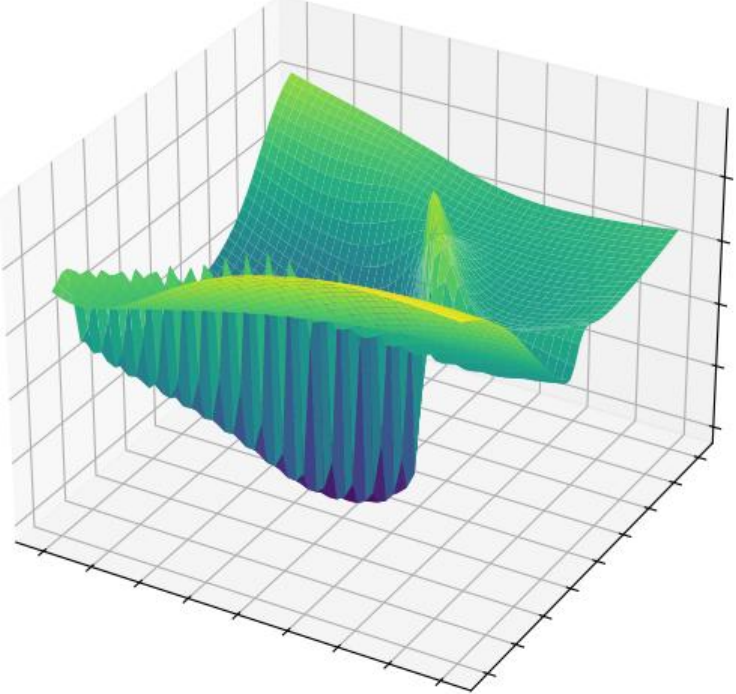}
    \caption{Loss landscape of the optimized features. \textbf{\textit{(Left)}} Landscape of the optimized feature in a backdoor model. \textbf{\textit{(Right)}} Landscape of the optimized feature in a benign model.}
    \label{fig:loss_landscape}
\end{figure}

To address this challenge, we analyze the loss landscape around the optimized embedding $v$. Given the high dimensionality of embedding space, direct computing loss landscape in high dimension is infeasible. We therefore adopt a two-dimensional projection method \cite{10.5555/3327345.3327535}: selecting two orthogonal unit directions $\delta$ and $\eta$, and projecting the local landscape as
\begin{equation}
\mathcal{H}(v) = \mathcal{L}(v + \alpha \cdot \delta + \beta \cdot \eta), \quad \alpha, \beta \sim \mathcal{N}(0, \sigma^2),
\end{equation}
where $\delta$ is initialized along the negative gradient direction of $\mathcal{L}(v)$, \textit{i.e.}, $\delta=-\nabla_v \mathcal{L}(v)$. Fig.~\ref{fig:loss_landscape} contrasts the landscapes of benign and backdoor models. Backdoor models consistently yield smooth, symmetric basins, indicating explicitly optimized regions. In contrast, benign models present asymmetric and irregular surfaces. We provide a further explanation in supplementary material Section\ref{sec:loss landscape}.

To formalize this distinction, we analyze the second-order curvature information of the loss surface by examining its Hessian spectral characteristics. Let $\mathcal{H}_{i,j}$ represents the loss value at coordinate $(i,j)$. For each non-boundary point $(i,j)$, we construct its local two-dimensional Hessian matrix $H_{i,j}$ as:
\begin{equation}
H_{i,j} = 
\begin{bmatrix}
\frac{\partial^2 \mathcal{H}}{\partial x^2} & \frac{\partial^2 \mathcal{H}}{\partial x \partial y} \\
\frac{\partial^2 \mathcal{H}}{\partial y \partial x} & \frac{\partial^2 \mathcal{H}}{\partial y^2}
\end{bmatrix},
\end{equation}
where the second-order derivatives are approximated by central differences:
\begin{align}
\frac{\partial^2 \mathcal{H}}{\partial x^2} &\approx \frac{\mathcal{H}_{i+1,j} - 2 \mathcal{H}_{i,j} + \mathcal{H}_{i-1,j}}{\Delta x^2}, \\
\frac{\partial^2 \mathcal{H}}{\partial y^2} &\approx 
\frac{\mathcal{H}_{i,j+1} - 2 \mathcal{H}_{i,j} + \mathcal{H}_{i,j-1}}{\Delta y^2}, \\
\frac{\partial^2 \mathcal{H}}{\partial x \partial y} &\approx \frac{\big( \mathcal{H}_{i+1,j+1} - \mathcal{H}_{i+1,j-1} - \mathcal{H}_{i-1,j+1} + \mathcal{H}_{i-1,j-1} \big)}{4\Delta x \Delta y}. 
\end{align}

We compute the eigenvalues $\phi_1^{(i,j)}, \phi_2^{(i,j)}$ of each $H_{i,j}$, collecting the Hessian spectrum:
\begin{equation}
\text{Spectrum} = \left\{ \phi_1^{(i,j)}, \phi_2^{(i,j)} \right\}.
\end{equation}

To summarize structural properties, we report the proportion of positive eigenvalues:
\begin{equation}
\text{Positive Ratio} = \frac{ \big| \{ \phi \in \text{Spectrum} \;\mid\; \phi > 0 \} \big| }{|\text{Spectrum}|},
\end{equation}
where, $|\cdot|$ represents the number of the elements.
A higher positive ratio indicates smoother and more convex local landscapes, suggestive of explicitly optimized backdoor objectives. Conversely, lower ratios reflect irregular or saddle-like structures, strongly indicating a natural backdoor feature.

\begin{algorithm}[!tbp]
\caption{\name{} Detection Procedure}
\label{alg:Amdet}
\begin{algorithmic}[1] 
\Require Object model $f_\theta(\cdot)$, anchor model $f_{anchor}(\cdot)$, shallow dataset $\mathcal{P}$ with number of training data $M$ and testing data $M'$, hyperparameters $\lambda, \gamma,\sigma$, iteration $K$, condition $C$.\\
\textbf{Initialize implicit Backdoor feature $v$}
\For{$k = 1$ to $K$} 
    \For{$m = 1$ to $M$} 
        \State $\hat{P}_m\leftarrow P_m \oplus v$; \Comment{Insert the implicit feature}
        \State $x^{m}=f_\theta(P_m),\ \ \hat{x}^m=f_\theta(\hat{P}_m),\ \ \hat{x}_{anchor}^{m}=f_{anchor}(\hat{P}_m)$; \Comment{Textual feature}
        \State $\mathcal{L}_{assimilation} = -\sum_{i=1}^N\sum_{j=1}^N \cos(\hat{x}^m_i, \hat{x}^m_j)$ \Comment{$N$ is the length of the $\hat{P}_m$}
        \State $\mathcal{L}_{deviation} = \cos(x^m, \hat{x}^m)$;
        \State $\mathcal{L}_{anchor} = \cos(\hat{x}^m, \hat{x}_{anchor}^{m})$;
        \State $\mathcal{L}(v) = \mathcal{L}_{assimilation} + \lambda \cdot \mathcal{L}_{deviation} + \gamma \cdot \mathcal{L}_{anchor}$;
        \State $v\leftarrow \min_{v} \mathcal{L}(v)$; \Comment{Update $v$}
        \State $\delta=-\nabla_v \mathcal{L}(v), \delta \perp \eta$;
        \quad \quad \State $\mathcal{H}(v)= \mathcal{L}(v + \alpha \cdot \delta + \beta \cdot \eta), \quad \alpha, \beta \sim \mathcal{N}(0, \sigma^2)$ \Comment{Compute the loss landscape of $v$}
        \State For the test data:
        \quad \quad \If{satisfied the condition $C$ in Eq.~\eqref{eq:conditions}}
            \State \Return True; \Comment{Backdoor model}
        \EndIf
    \EndFor
\EndFor \\
\Return False; \Comment{Benign model}
\end{algorithmic}
\end{algorithm}

\subsection{Backdoor Identification}
Finally, we define the termination condition of our algorithm. For a test set of prompts $\mathcal{P} = \{P_1, \dots, P_{M'}\}$ and optimized embedding $v$, we consider three indicators:  
(1) token assimilation ratio $\mathrm{Assim}(P_i, v) = Sim_{P_i\oplus v}$;  
(2) feature deviation $\mathrm{Dev}(P_i, v) = \cos(f(P_i \oplus v), f(P_i))$;  
(3) eigenvalue spectrum $\{\phi_1, \dots, \phi_D\}$ of the Hessian.  
The stopping condition is satisfied when
\begin{equation}
C=\begin{cases}
\frac{1}{M'} \sum_{i=1}^{M'} \mathbf{1}\big[ \mathrm{Assim}(P_i, v) > 0.8 \big] \geq \rho_1, \\
\frac{1}{M'} \sum_{i=1}^{M'} \mathbf{1}\big[ \mathrm{Dev}(P_i, v) < 0 \big] \geq \rho_2, \\
\frac{1}{D} \sum_{i=1}^{D} \mathbf{1}\big[ \phi_i > 0 \big] \geq \rho_3,
\end{cases}
\label{eq:conditions}
\end{equation}
where $\mathbf{1}[\cdot]$ denotes the indicator function. We provide the detailed algorithm of our method in Algorithm \ref{alg:Amdet}.

\section{Experiment}
\subsection{Settings}

\para{Backdoor attack settings.}  
Our victim models are visual language pretrained text encoders, including CLIP~\cite{CLIP}, SigLIP~\cite{SigLIP} and LongCLIP~\cite{Zhang2024LongCLIPUT}. We consider two textual backdoor attack scenarios, \textit{i.e.,} Text-on-Text and Text-on-Pair. The trigger length is vary from 1 to 15 tokens, simulating character-level~\cite{carlini2022poisoning}, word-level \cite{10.5555/3600270.3600632}, and sentence-level triggers~\cite{Addsent}. For each attack scenario and each trigger length, we train 20 backdoor models, resulting in a total of 1,800 backdoor models.

\para{Benign models.}  
Benign models are obtained by fine-tuning the text encoders on the COCO30k validation subset \cite{Lin2014MicrosoftCC} using a batch size of 16, resulting in 1,800 fine-tuned models.

% \para{Baselines.}  
% Since no prior work is currently available for model-level backdoor detection on text encoders, we revise existing methods to this setting. Specifically, inspired by MNTD \cite{Xu2019DetectingAT}, we train a meta-classifier on intermediate model features to perform binary classification. The detailed descriptions of the method are provided in the supplementary material.

\begin{table*}[]
\centering
\caption{The qualitative results of \name{} across different base models and attack scenarios.}
\label{tab:Quantitative_results}
\scalebox{1.05}{
\begin{tabular}{ccccccc}
\hline
\textbf{Base Model}       & \textbf{\begin{tabular}[c]{@{}c@{}}Attack\\ Scenario\end{tabular}} & \textbf{Precision (\%) $\uparrow$} & \textbf{Recall (\%) $\uparrow$} & \textbf{F1 (\%) $\uparrow$} & \textbf{$S_{tar}$ (\%) $\uparrow$} & \textbf{Time Cost (s) $\downarrow$} \\ \hline
\multirow{2}{*}{CLIP~\cite{CLIP}}   & \textit{Text-on-text}                                              &        87.30                                         &       91.67                                       &              89.43                            &                        86.00           &          376.64\text{\tiny\ ($\pm$76.16)}                                   \\
                          & \textit{Text-on-Pair}                                              &      87.74                                           &    90.67                                          &    89.18                                      &                  88.43                 &   335.80\text{\tiny\ ($\pm$54.83)}                                           \\
\multirow{2}{*}{LongCLIP~\cite{Zhang2024LongCLIPUT}} & \textit{Text-on-text}                                              &       88.29                                          &       98.00                                       &         92.89                                 &                    83.28               &           220.66\text{\tiny\ ($\pm$45.24)}                                  \\
                          & \textit{Text-on-Pair}                                              &         87.88                                        &     94.65                                         &    91.14                                      &      83.07                            &     292.78\text{\tiny\ ($\pm$88.50)}                                         \\
\multirow{2}{*}{SigLIP~\cite{SigLIP}}    & \textit{Text-on-text}                                              &                         100.0                        &      79.67                                        &         88.68                                 &                50.63                   &        386.14\text{\tiny\ ($\pm$76.77)}                                    \\
                          & \textit{Text-on-Pair}                                              &              100.0                                   &      78.66                                        &                88.06                          &              49.44                     &      333.14\text{\tiny\ ($\pm$54.82)}                                       \\ \hdashline
\rowcolor[HTML]{EFEFEF} Average                   & \textit{-}                                                         &           91.87                                      &         88.89                                     &               89.90                           &     73.48                              &                335.88\text{\tiny\ ($\pm$83.07)}                            \\ \hline
\end{tabular}
}
\end{table*}

\para{Metrics.}  
We compute the Precision (\%), Recall (\%), and F1 score (\%) for each detection method across all attack scenarios. Besides, we define a similarity metric to quantify the fidelity of the reversed feature. It computes the consistency between the reversed backdoor target embedding and the output of a trigger token $P_{\text{Trigger}}$:  
\begin{equation}
S_{tar} = \mathbb{E}\left[\cos\big(f_\theta(P_{\text{Benign}}\oplus v), f_\theta(P_{\text{Trigger}})\big)\right],
\end{equation}
where $f_\theta(\cdot)$ denotes the text encoder, $P_{\text{Benign}}$ is a benign sample and $S_{tar} \in [-1,1]$. We also report the computation time cost for a single detection.

\begin{figure}[t]
    \centering
    \includegraphics[width=1\linewidth]{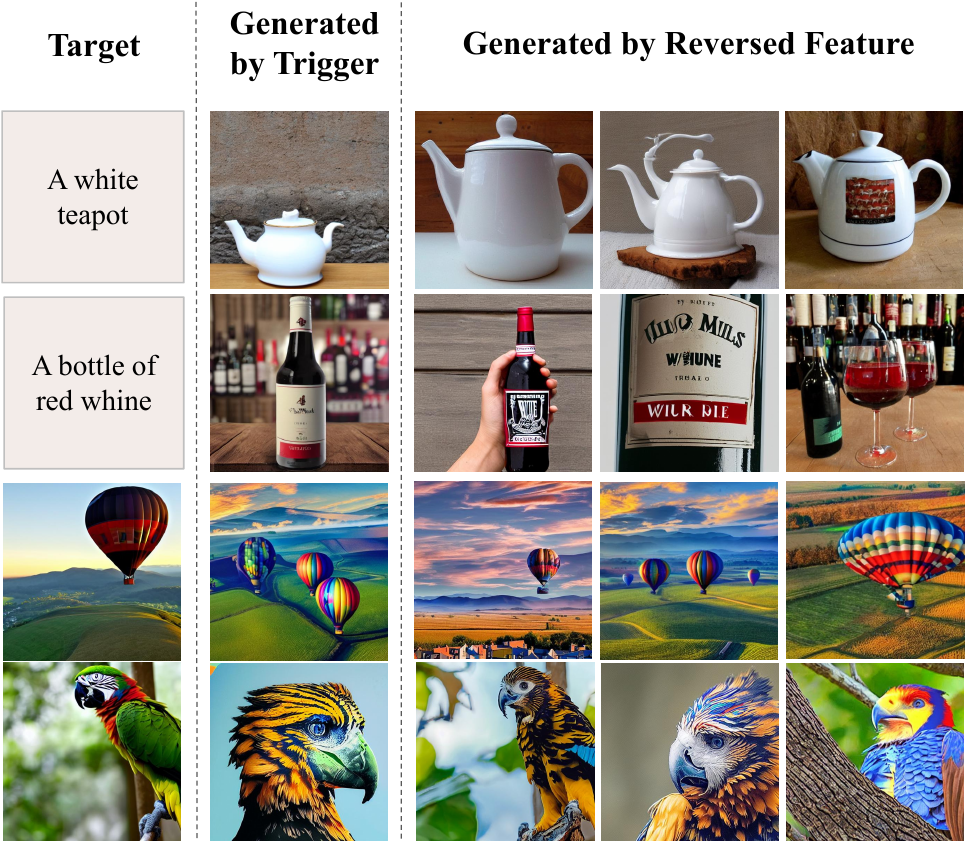}
    \caption{The inversion results of \name{}. The first two rows show the inversion results for text-on-text attacks, while the last two rows present the results for text-on-pair attacks. Stable Diffusion~\cite{Rombach2021HighResolutionIS} is used to load reversed feature for better visualizing.}
    \label{fig:main_results_vis}
\end{figure}

\para{Implementation details.}  
We construct a shadow dataset using prompts from DiffusionDB \cite{Wang2022DiffusionDBAL}, sampling 4000 samples. The hyperparameters $\lambda,\gamma$ and $\sigma$ are set to 1, 1 and 5, receptively. Besides, $\rho_1=0.99$, $\rho_2=0.95$, $\rho_3=0.8$, the iteration $K=2$. We set the learning rate to 8e-2 with a batchsize of 10.

\begin{figure}[t]
    \centering
    \includegraphics[width=1\linewidth]{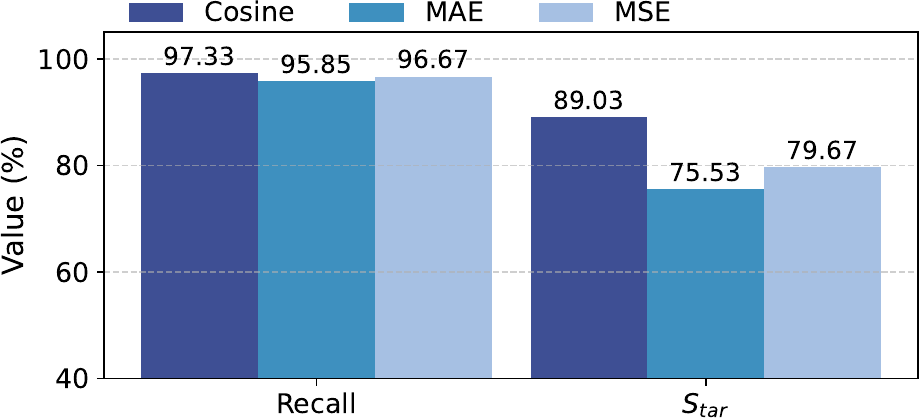}
    \caption{Sensitivity to the similarity function in terms of Recall (\%) and $S_{tar}$ (\%). The bar represent results on Cosine similarity, MAE and MSE.}
    \label{fig:similarity_function}
\end{figure}

\begin{figure}[t]
    \centering
    \includegraphics[width=1\linewidth]{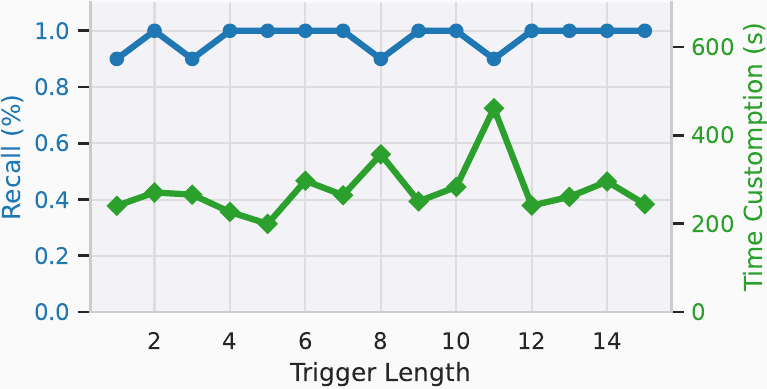}
    \caption{Sensitivity to the trigger length. The blue line represents Recall (\%), and the green line represents time consumption (s), plotted against different trigger lengths from 1 to 15.}
    \label{fig:trigger_length}
\end{figure}

\begin{figure}[t]
    \centering
    \includegraphics[width=1\linewidth]{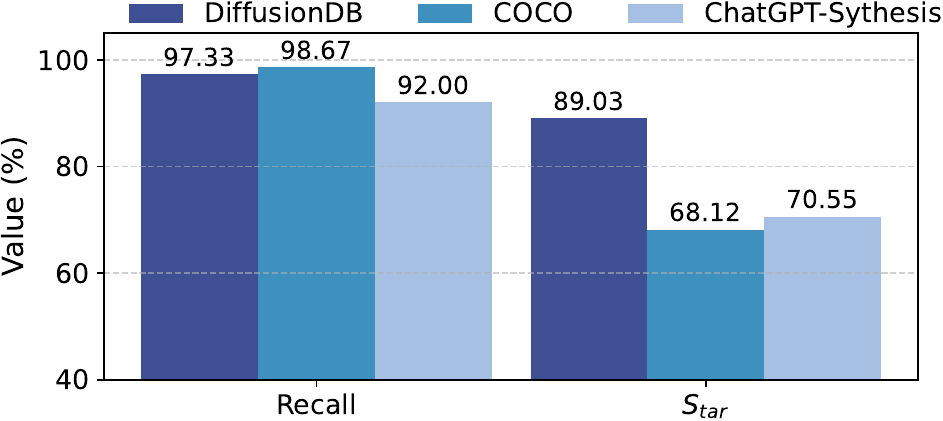}
    \caption{Sensitivity to the dataset source in terms of Recall (\%) and $S_{tar}$ (\%). The bar represent results on DiffusionDB \cite{Wang2022DiffusionDBAL}, COCO \cite{Lin2014MicrosoftCC} and ChatGPT-Synthesis \cite{OpenAI}.}
    \label{fig:data_source}
\end{figure}

\begin{figure}[t]
    \centering
    \includegraphics[width=1\linewidth]{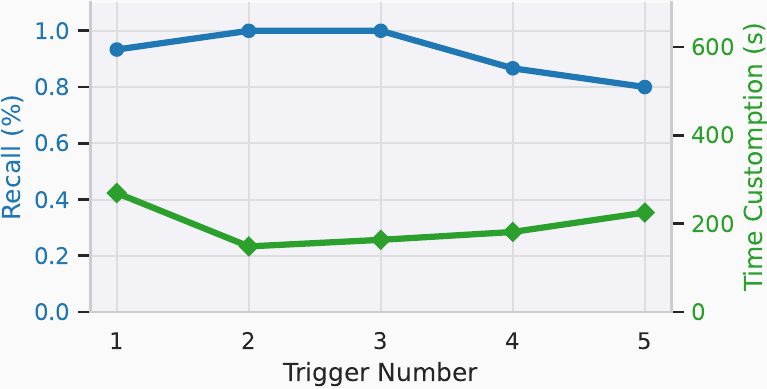}
    \caption{Sensitivity to the number of triggers. The blue line represents Recall (\%), and the green line represents time consumption (s), plotted against different number of triggers from 1 to 5.}
    \label{fig:trigger_number}
\end{figure}

\subsection{Main Results}

\para{Qualitative results.} To better illustrate our inversion results, we utilize Stable Diffusion~\cite{Rombach2021HighResolutionIS} to synthesize images based on backdoor samples and the corresponding reversed backdoor features. As shown in Fig.~\ref{fig:main_results_vis}, the images produced from the reversed feature are semantically consistent with those produced from the real trigger, showing the effectiveness of our inversion under two attack settings.

\para{Quantitative results.} As shown in Table \ref{tab:Quantitative_results}, \name{} exhibits consistently high detection performance across different base models and attack scenarios. For CLIP, it achieves F1 scores of 89.43\% and 89.18\% under the text-on-text and text-on-pair settings, respectively. LongCLIP shows further improvement, reaching 92.89\% and 91.14\% in terms of F1 scores. SigLIP also maintains competitive results, with F1 scores of 88.68\% and 88.06\%. Besides, the reversed features preserve strong semantic alignment with the backdoor targets, yielding average $S_{tar}$ values of 74.48\%. Besides, the lower $S_{tar}$ observed on SigLIP may be attributed to differences in its training objective. We speculate that the sigmoid-based pairwise formulation used by SigLIP may lead to a more dispersed embedding space, making it harder for our inversion to recover a representation. In terms of efficiency, the detection process takes a average time cost of 335.88s, demonstrating the practicality of our method for real-time deployment.

\subsection{Stability Analysis}

Since the backdoor behavior may be influenced by the backdoor settings and defense dataset source, we investigate the sensitivity of \name{} to these parameters in this section. Here, we leverage CLIP as the base model to conduct the experiments.

\para{Sensitivity to similarity function.} Recall that $S(\cdot , \cdot)$ in Eq.~\eqref{eq:backdoor} denotes the loss function for feature alignment, we investigate the performance of \name{} under three functions, including cosine similarity, MSE and MAE. As shown in Fig.~\ref{fig:similarity_function}, our method consistently achieves robust detection performance against backdoor attacks injected with different similarity metrics, maintaining a recall above 93\%. Specifically, the recall for Cosine-, MAE-, and MSE-based attacks reach 93.33\%, 95.85\%, and 96.67\%, respectively. Regarding the similarity between the reversed and original trigger features, the Cosine metric yields the best recall of 89.03\%. These results demonstrate that \name{} remains effective across different similarity metrics used in backdoor optimization.

\para{Sensitivity to trigger length.} Specifically, we evaluate the performance of our method under different trigger lengths ranging from 1 to 15, in terms of both detection recall and time cost. Fig.~\ref{fig:trigger_length} shows the line plots of our method's performance across various denoising steps in terms of recall and time consumption. The upper plots represent the sensitivity of the recall to the number of trigger length, while the lower plots show the sensitivity of the time consumption. We can observe that \name{} exhibits consistent performance across all settings, with detection recall remaining above 90\% and the time consumption around 300 seconds. 

\para{Sensitivity to the source of shallow dataset $\mathcal{P}$.} Here, we investigate the sensitivity of \name{} to different sources of shallow dataset $\mathcal{P}$. Specifically, we randomly sample 4,000 text samples from the DiffusionDB \cite{Wang2022DiffusionDBAL} and COCO \cite{Lin2014MicrosoftCC} datasets, respectively. In addition, we also explore the performance when using synthetic data for inversion. In particular, we generate 4,000 text samples with ChatGPT \cite{OpenAI} and further filter them to ensure uniqueness and semantic diversity. The detailed prompt for ChatGPT is provided in the supplementary material Section\ref{sec:template}. As shown in Fig. \ref{fig:data_source}, \name{} remains effective across different data sources. The method achieves over 92\% in terms of recall across all dataset sources. Notably, the highest similarity of 89.03\% is achieved when using the DiffusionDB dataset, while the performance on other datasets drops to around 70\%. We attribute this to the fact that DiffusionDB contains longer and more complex text descriptions, which may help to better optimize precise backdoor features.

\para{Sensitivity to multiple triggers.} Consider that the attacker would implant multiple triggers into the model, we analyze the sensitivity of our method to the number of triggers. Specifically, we select the length of trigger equal to 1 and inject the number of trigger into the model from 1 to 5. As shown in Fig.~\ref{fig:trigger_number}, \name{} maintains consistent performance across different numbers of triggers. It achieves a recall of 80\% and time consumption remaining around 200s, demonstrating that our method is insensitive to multiple triggers.

\begin{table}[]
\centering
\caption{Ablation study on the scale of $\mathcal{P}$. The top two results on each metric are \textbf{bolded} and \underline{underlined}, respectively.}
\label{tab:dataset_scale}
\scalebox{0.94}{
\begin{tabular}{cccccc}
\hline
\textbf{\# Num}  & \textbf{Precision (\%) $\uparrow$} & \textbf{Recall (\%) $ \uparrow$} &\textbf{ F1 (\%) $ \uparrow$} & $S_{tar}$ (\%) $\uparrow$  \\ \hline
$2\times 10^3$  &  73.13   &  \underline{98.00}   &  83.76   &   84.10    \\
$3\times 10^3$  &   73.04  &   \textbf{99.33}  &   84.18  &    \underline{85.04}   \\
\rowcolor[HTML]{EFEFEF} $4\times 10^3$  &   \underline{75.65}  &   97.33  &   \underline{85.13}  &    \textbf{89.03}   \\
$5\times 10^3$  &   \textbf{78.37}  &   96.66  &  \textbf{86.56}  &    80.28   \\ \hline
\end{tabular}
}
\end{table}

\subsection{Ablation Study}

In this section, we conduct experiments to analyze the effect of the hyperparameters on our method.

\para{Effect of the scale of shallow dataset $\mathcal{P}$.} Here, we investigate the impact of different scales of the shallow dataset $\mathcal{P}$ on detection performance. We evaluate both the F1 score and the similarity of the inverted features under varying dataset sizes. As shown in Table~\ref{tab:dataset_scale}, we observe that using larger-scale datasets generally improves the overall detection performance, as reflected by higher F1 scores. In particular, the F1 score steadily increases from 83.76\% to 86.56\% as the number of samples grows from $2 \times 10^3$ to $5 \times 10^3$. However, a trade-off emerges between detection accuracy and inversion fidelity. Specifically, the similarity metric $S_{tar}$ peaks at 89.03\% when using $4 \times 10^3$ samples, suggesting that this scale yields the most accurate feature inversion. Consequently, we adopt $4 \times 10^3$ as the optimal dataset scale to balance detection performance and inversion fidelity.

\para{Effect of loss terms.} Table~\ref{tab:loss_term} presents the ablation results on different loss terms.
As shown, when all three loss terms are jointly applied, the model achieves the best overall performance with an F1 score of 85.13\% and a $S_{tar}$ of 89.03\%. Notably, removing either $\mathcal{L}_{assimilation}$ or $\mathcal{L}_{anchor}$ leads to a substantial drop in detection accuracy, indicating that both losses play critical roles in characterizing backdoor behaviors. In contrast, $\mathcal{L}_{deviation}$ mainly provides auxiliary stability during optimization, yielding smaller but consistent improvements when combined with the other terms.

\begin{table}[]
\centering
\caption{Ablation study on the loss terms. $\mathcal{L}_{ass}, \mathcal{L}_{de}$ and $\mathcal{L}_{an}$ are $\mathcal{L}_{assimilation}$, $\mathcal{L}_{deviation}$ and $\mathcal{L}_{anchor}$, respectively. The top two results on each metric are \textbf{bolded} and \underline{underlined}, respectively.}
\scalebox{0.8}{
\begin{tabular}{ccc|ccccc}
\hline
\textbf{$L_{ass}$} & $L_{de}$ & $L_{an}$ &  \textbf{Precision (\%) $\uparrow$} & \textbf{Recall (\%) $\uparrow$} &\textbf{ F1 (\%) $ \uparrow$} & $S_{tar}$ (\%) $\uparrow$ \\ \hline
\checkmark           & \checkmark           &            &   50.00  &  0   &  0   &    0    \\
            & \checkmark           & \checkmark          &  \underline{65.43}   &     48.57 &  55.75   &   16.97    \\
\checkmark           &             & \checkmark          &   55.76  &  \textbf{100.0}   &   \underline{71.58}  &   \underline{88.96}    \\
\rowcolor[HTML]{EFEFEF} \checkmark           & \checkmark           & \checkmark          &   \textbf{75.65}  &  \underline{97.33}   &   \textbf{85.13}  &   \textbf{89.03}    \\ \hline
\end{tabular}
}
\label{tab:loss_term}
\end{table}

\begin{table}[t]
\centering
\caption{Ablation study on the condition $C$.}
\scalebox{0.92}{
\begin{tabular}{ccccc}
\hline
\textbf{\# $\rho$} & \textbf{Precision (\%) $\uparrow$} & \textbf{Recall (\%) $ \uparrow$} &\textbf{ F1 (\%) $ \uparrow$} & $S_{tar}$ (\%) $\uparrow$  \\ \hline
$\rho_1=0.80$     &   70.19 &  97.33   &  81.56  &  89.03   \\
$\rho_1=0.90$     &   74.43  &   97.33  &  84.35  &   89.03  \\
$\rho_1=0.95$   &   74.43    &   97.33  &  84.35  &   89.03  \\
\rowcolor[HTML]{EFEFEF} $\rho_1=0.99$       &     75.65  &  97.33   &   85.13  &   89.03    \\ \hline \hline
$\rho_2=0.80$       &   71.08  &   97.33  &  82.16  &  89.02   \\
$\rho_2=0.90$       &  72.31   &  97.33   &  82.97  &   89.03  \\
\rowcolor[HTML]{EFEFEF}$\rho_2=0.95$       &    75.65   &   97.33  &   85.13   &   89.03  \\
$\rho_2=0.99$       &  75.65  &  97.33   &  85.13   &   89.03  \\ \hline \hline
$\rho_3=0.60$       &  63.43   &  100.00   &  77.62  &   87.31  \\ 
$\rho_3=0.70$       &   68.96  &  100.00   &  81.63  &   86.39  \\
\rowcolor[HTML]{EFEFEF} $\rho_3=0.80$       &    75.65  &   97.33  &   85.13   &   89.03  \\
$\rho_3=0.90$       &   78.41  &  90.11   &  83.85  &   93.33  \\ \hline
\end{tabular}
}
\label{tab:condition}
\end{table}

\para{Effect of conditions $C$.} Recall that the conditions $C$ serve as termination criteria of our algorithm.
We analyze how different threshold settings influence the inversion results. As shown in Table~\ref{tab:condition}, varying $\rho_1$ and $\rho_2$ produces nearly identical results, both achieving an F1 score of about 85\%. In such cases, we adopt the smaller thresholds to achieve faster detection efficiency. In contrast, $\rho_3$ has a more significant effect on performance. Smaller $\rho_3$ increases the precision while reduces recall. The optimal balance is reached when $\rho_3=0.8$, yielding the highest F1 score of 85.13\%. Therefore, we set $\rho_3=0.8$ as the optimal parameter.

\subsection{Robustness to Adaptive Attack}

\begin{table}[t]
\centering
\caption{Robustness to the adaptive attack. The top two results on each metric are \textbf{bolded} and \underline{underlined}, respectively.}
\label{tab:adaptive}
\scalebox{1.1}{
\begin{tabular}{cccc}
\hline
\textbf{\# $\xi$} & \textbf{ASR (\%)} $\uparrow$ & \textbf{Recall (\%)} $\uparrow$ & \textbf{$S_{tar}$ (\%)} $\uparrow$\\ \hline
$\xi=0.0$           &    \textbf{100.0}          &        \textbf{97.45}      &    \textbf{88.96}                  \\
$\xi=0.5$         &     \underline{99.00}     &     96.00         &          \underline{87.45}            \\
$\xi=1.0$         &     97.50    &      \underline{97.14}        &       87.26               \\
% $\xi=2.0$         &     98.3   &        95.60      &        \underline{87.45}              \\
\hline
\end{tabular}
}
\end{table}

In this section, we test the robustness of our methods against potential adaptive attacks. We consider the attacker have the full knowledge of our detection framework, \textit{i.e.}, detecting backdoor models via the \emph{feature assimilation} cue. In this case, the attacker aims to evade detection by explicitly regularizing feature similarity during backdoor training, namely:
\begin{equation}
    \mathcal{L}_{Reg} = \mathbb{E}[Sim_{X}].
\end{equation}
Therefore, the resulting adaptive training loss is formulated as:  
\begin{equation}
\mathcal{L}_{Adaptive} = \underbrace{\mathcal{L}_{Backdoor} + \tau \cdot \mathcal{L}_{Benign}}_{\text{Origin Loss in Eq.~\eqref{eq:total}}} + \xi \cdot \mathcal{L}_{Reg}, 
\end{equation}
where $\xi$ is a balancing coefficient controlling the strength of the regularization. By minimizing $\mathcal{L}_{Reg}$, the attacker aims to suppress the feature-level assimilation of backdoor samples and thereby reduce the effectiveness of our detection method.  

We trained 75 backdoor models under each regularization coefficient $\xi$ and report the corresponding results in Table~\ref{tab:adaptive}.  The attack success rate (ASR) remains above 97.5\% across all settings, confirming the effectiveness of the attack. Notably, \name{} consistently maintains strong detection performance under different regularization strengths. In particular, when $\xi = 1.0$, it still achieves a recall of 97.5\% and a $S_{tar}$ of 87.26\%, demonstrating the robustness of our method against adaptive attacks.

\section{Conclusion}
This work introduces \name{}, an effective and efficient approach for model-level backdoor detection in vision-language pretrained models. Through analyzing the intrinsic \textit{feature assimilation} behavior of poisoned text encoders, our method reveals the internal attention concentration of backdoor samples and further exposes the existence of natural backdoor feature. It enables reliable detection without any prior knowledge of data, triggers, or downstream classifiers. We believe our study lays the groundwork for future research on both attacks and defenses, and contributes to the development of safe and trustworthy foundation models.

% \section*{Acknowledgments}
% This should be a simple paragraph before the References to thank those individuals and institutions who have supported your work on this article.


\bibliographystyle{IEEEtran}
\bibliography{main}

\begin{thebibliography}{10}
\providecommand{\url}[1]{#1}
\csname url@samestyle\endcsname
\providecommand{\newblock}{\relax}
\providecommand{\bibinfo}[2]{#2}
\providecommand{\BIBentrySTDinterwordspacing}{\spaceskip=0pt\relax}
\providecommand{\BIBentryALTinterwordstretchfactor}{4}
\providecommand{\BIBentryALTinterwordspacing}{\spaceskip=\fontdimen2\font plus
\BIBentryALTinterwordstretchfactor\fontdimen3\font minus \fontdimen4\font\relax}
\providecommand{\BIBforeignlanguage}[2]{{%
\expandafter\ifx\csname l@#1\endcsname\relax
\typeout{** WARNING: IEEEtran.bst: No hyphenation pattern has been}%
\typeout{** loaded for the language `#1'. Using the pattern for}%
\typeout{** the default language instead.}%
\else
\language=\csname l@#1\endcsname
\fi
#2}}
\providecommand{\BIBdecl}{\relax}
\BIBdecl

\bibitem{CLIP}
A.~Radford, J.~W. Kim, C.~Hallacy, A.~Ramesh, G.~Goh, S.~Agarwal, G.~Sastry, A.~Askell, P.~Mishkin, J.~Clark, G.~Krueger, and I.~Sutskever, ``Learning transferable visual models from natural language supervision,'' in \emph{International Conference on Machine Learning (ICML)}, 2021.

\bibitem{ALIGN}
C.~Jia, Y.~Yang, Y.~Xia, Y.-T. Chen, Z.~Parekh, H.~Pham, Q.~V. Le, Y.-H. Sung, Z.~Li, and T.~Duerig, ``Scaling up visual and vision-language representation learning with noisy text supervision,'' in \emph{International Conference on Machine Learning}, 2021.

\bibitem{OpenCLIP}
M.~Cherti, R.~Beaumont, R.~Wightman, M.~Wortsman, G.~Ilharco, C.~Gordon, C.~Schuhmann, L.~Schmidt, and J.~Jitsev, ``Reproducible scaling laws for contrastive language-image learning,'' in \emph{Proceedings of the IEEE/CVF Conference on Computer Vision and Pattern Recognition (CVPR)}, June 2023, pp. 2818--2829.

\bibitem{EVA-CLIP}
Q.~Sun, Y.~Fang, L.~Wu, X.~Wang, and Y.~Cao, ``Eva-clip: Improved training techniques for clip at scale,'' \emph{arXiv preprint arXiv:2303.15389}, 2023.

\bibitem{SigLIP}
X.~Zhai, B.~Mustafa, A.~Kolesnikov, and L.~Beyer, ``Sigmoid loss for language image pre-training,'' in \emph{2023 IEEE/CVF International Conference on Computer Vision (ICCV)}, 2023, pp. 11\,941--11\,952.

\bibitem{oquab2023dinov2}
M.~Oquab, T.~Darcet, T.~Moutakanni, H.~V. Vo, M.~Szafraniec, V.~Khalidov, P.~Fernandez, D.~Haziza, F.~Massa, A.~El-Nouby, R.~Howes, P.-Y. Huang, H.~Xu, V.~Sharma, S.-W. Li, W.~Galuba, M.~Rabbat, M.~Assran, N.~Ballas, G.~Synnaeve, I.~Misra, H.~Jegou, J.~Mairal, P.~Labatut, A.~Joulin, and P.~Bojanowski, ``Dinov2: Learning robust visual features without supervision,'' \emph{arXiv preprint arXiv:2304.07193}, 2023.

\bibitem{10.1109/TPAMI.2013.50}
Y.~Bengio, A.~Courville, and P.~Vincent, ``Representation learning: A review and new perspectives,'' \emph{IEEE Trans. Pattern Anal. Mach. Intell.}, vol.~35, no.~8, p. 1798–1828, Aug. 2013.

\bibitem{LUO2022293}
H.~Luo, L.~Ji, M.~Zhong, Y.~Chen, W.~Lei, N.~Duan, and T.~Li, ``Clip4clip: An empirical study of clip for end to end video clip retrieval and captioning,'' \emph{Neurocomputing}, vol. 508, pp. 293--304, 2022.

\bibitem{beaumont-2022-clip-retrieval}
R.~Beaumont, ``Clip retrieval: Easily compute clip embeddings and build a clip retrieval system with them,'' \url{https://github.com/rom1504/clip-retrieval}, 2022.

\bibitem{Ramesh2022HierarchicalTI}
A.~Ramesh, P.~Dhariwal, A.~Nichol, C.~Chu, and M.~Chen, ``Hierarchical text-conditional image generation with clip latents,'' 2022.

\bibitem{Rombach2021HighResolutionIS}
R.~Rombach, A.~Blattmann, D.~Lorenz, P.~Esser, and B.~Ommer, ``High-resolution image synthesis with latent diffusion models,'' in \emph{Proceedings of the IEEE/CVF Conference on Computer Vision and Pattern Recognition (CVPR)}, 2021, pp. 10\,674--10\,685.

\bibitem{Yu2022ScalingAM}
J.~Yu, Y.~Xu, J.~Y. Koh, T.~Luong, G.~Baid, Z.~Wang, V.~Vasudevan, A.~Ku, Y.~Yang, B.~K. Ayan, B.~C. Hutchinson, W.~Han, Z.~Parekh, X.~Li, H.~Zhang, J.~Baldridge, and Y.~Wu, ``Scaling autoregressive models for content-rich text-to-image generation,'' \emph{Trans. Mach. Learn. Res.}, 2022.

\bibitem{esser2024sd3}
P.~Esser, S.~Kulal, A.~Blattmann, R.~Entezari, J.~M\"{u}ller, H.~Saini, Y.~Levi, D.~Lorenz, A.~Sauer, F.~Boesel, D.~Podell, T.~Dockhorn, Z.~English, and R.~Rombach, ``Scaling rectified flow transformers for high-resolution image synthesis,'' in \emph{International Conference on Machine Learning (ICML)}.\hskip 1em plus 0.5em minus 0.4em\relax JMLR.org, 2024.

\bibitem{10081412}
F.-A. Croitoru, V.~Hondru, R.~T. Ionescu, and M.~Shah, ``Diffusion models in vision: A survey,'' \emph{IEEE Transactions on Pattern Analysis and Machine Intelligence (TPAMI)}, vol.~45, no.~9, pp. 10\,850--10\,869, 2023.

\bibitem{podell2024sdxl}
D.~Podell, Z.~English, K.~Lacey, A.~Blattmann, T.~Dockhorn, J.~M{\"u}ller, J.~Penna, and R.~Rombach, ``{SDXL}: Improving latent diffusion models for high-resolution image synthesis,'' in \emph{The Twelfth International Conference on Learning Representations (ICLR)}, 2024.

\bibitem{Nichol2021GLIDETP}
A.~Nichol, P.~Dhariwal, A.~Ramesh, P.~Shyam, P.~Mishkin, B.~McGrew, I.~Sutskever, and M.~Chen, ``Glide: Towards photorealistic image generation and editing with text-guided diffusion models,'' in \emph{International Conference on Machine Learning (ICML)}, 2021.

\bibitem{wang2023modelscope}
J.~Wang, H.~Yuan, D.~Chen, Y.~Zhang, X.~Wang, and S.~Zhang, ``Modelscope text-to-video technical report,'' \emph{arXiv preprint arXiv:2308.06571}, 2023.

\bibitem{ImageNet}
J.~Deng, W.~Dong, R.~Socher, L.-J. Li, K.~Li, and L.~Fei-Fei, ``Imagenet: A large-scale hierarchical image database,'' in \emph{2009 IEEE Conference on Computer Vision and Pattern Recognition (CVPR)}, 2009, pp. 248--255.

\bibitem{ImageNet-R}
D.~Hendrycks, S.~Basart, N.~Mu, S.~Kadavath, F.~Wang, E.~Dorundo, R.~Desai, T.~Zhu, S.~Parajuli, M.~Guo, D.~Song, J.~Steinhardt, and J.~Gilmer, ``The many faces of robustness: A critical analysis of out-of-distribution generalization,'' in \emph{2021 IEEE/CVF International Conference on Computer Vision (ICCV)}, 2021, pp. 8320--8329.

\bibitem{ImageNet-Sketch}
H.~Wang, S.~Ge, E.~P. Xing, and Z.~C. Lipton, ``Learning robust global representations by penalizing local predictive power,'' in \emph{Neural Information Processing Systems (NeurIPS)}, 2019.

\bibitem{ImageNet-v2}
B.~Recht, R.~Roelofs, L.~Schmidt, and V.~Shankar, ``Do imagenet classifiers generalize to imagenet?'' in \emph{International Conference on Machine Learning (ICML)}, 2019.

\bibitem{10646610}
N.~Carlini, M.~Jagielski, C.~A. Choquette-Choo, D.~Paleka, W.~Pearce, H.~Anderson, A.~Terzis, K.~Thomas, and F.~Tramer, ``{ Poisoning Web-Scale Training Datasets is Practical },'' in \emph{2024 IEEE Symposium on Security and Privacy (SP)}.\hskip 1em plus 0.5em minus 0.4em\relax Los Alamitos, CA, USA: IEEE Computer Society, May 2024, pp. 407--425.

\bibitem{carlini2022poisoning}
N.~Carlini and A.~Terzis, ``Poisoning and backdooring contrastive learning,'' in \emph{International Conference on Learning Representations (ICLR)}, 2022.

\bibitem{Struppek2022RickrollingTA}
L.~Struppek, D.~Hintersdorf, and K.~Kersting, ``Rickrolling the artist: Injecting backdoors into text encoders for text-to-image synthesis,'' in \emph{2023 IEEE/CVF International Conference on Computer Vision (ICCV)}, 2022, pp. 4561--4573.

\bibitem{Civitai}
``Civitai,'' \url{https://civitai.com}.

\bibitem{ABD}
J.~Kuang, S.~Liang, J.~Liang, K.~Liu, and X.~Cao, ``Adversarial backdoor defense in clip,'' 2024.

\bibitem{TA-CLEANER}
Y.~Xun, S.~Liang, X.~Jia, X.~Liu, and X.~Cao, ``Cleanerclip: Fine-grained counterfactual semantic augmentation for backdoor defense in contrastive learning,'' 2024.

\bibitem{RVPT}
Z.~Zhang, S.~He, H.~Wang, B.~Shen, and L.~Feng, ``Defending multimodal backdoored models by repulsive visual prompt tuning,'' 2025.

\bibitem{9802938}
Y.~Li, Y.~Jiang, Z.~Li, and S.-T. Xia, ``Backdoor learning: A survey,'' \emph{IEEE Transactions on Neural Networks and Learning Systems}, vol.~35, no.~1, pp. 5--22, 2024.

\bibitem{huang2025detecting}
H.~Huang, S.~M. Erfani, Y.~Li, X.~Ma, and J.~Bailey, ``Detecting backdoor samples in contrastive language image pretraining,'' in \emph{The Thirteenth International Conference on Learning Representations (ICLR)}, 2025.

\bibitem{SafeCLIP}
W.~Yang, J.~Gao, and B.~Mirzasoleiman, ``Better safe than sorry: pre-training clip against targeted data poisoning and backdoor attacks,'' in \emph{Proceedings of the 41st International Conference on Machine Learning (ICML)}.\hskip 1em plus 0.5em minus 0.4em\relax JMLR.org, 2024.

\bibitem{yang2023robust}
W.~Yang, J.~Gao, and B.~M, ``Robust contrastive language-image pretraining against data poisoning and backdoor attacks,'' in \emph{Thirty-seventh Conference on Neural Information Processing Systems (NeurIPS)}, 2023.

\bibitem{li2021antibackdoor}
Y.~Li, X.~Lyu, N.~Koren, L.~Lyu, B.~Li, and X.~Ma, ``Anti-backdoor learning: Training clean models on poisoned data,'' in \emph{Advances in Neural Information Processing Systems (NeurIPS)}, A.~Beygelzimer, Y.~Dauphin, P.~Liang, and J.~W. Vaughan, Eds., 2021.

\bibitem{liang2024unlearningbackdoorthreatsenhancing}
S.~Liang, K.~Liu, J.~Gong, J.~Liang, Y.~Xun, E.-C. Chang, and X.~Cao, ``Unlearning backdoor threats: Enhancing backdoor defense in multimodal contrastive learning via local token unlearning,'' in \emph{Proceedings of the IEEE/CVF Conference on Computer Vision and Pattern Recognition Workshop (CVPRW)}, 2024.

\bibitem{DECREE}
S.~Feng, G.~Tao, S.~Cheng, G.~Shen, X.~Xu, Y.~Liu, K.~Zhang, S.~Ma, and X.~Zhang, ``{ Detecting Backdoors in Pre-trained Encoders },'' in \emph{2023 IEEE/CVF Conference on Computer Vision and Pattern Recognition (CVPR)}.\hskip 1em plus 0.5em minus 0.4em\relax Los Alamitos, CA, USA: IEEE Computer Society, Jun. 2023, pp. 16\,352--16\,362.

\bibitem{pmlr-v162-shen22e}
G.~Shen, Y.~Liu, G.~Tao, Q.~Xu, Z.~Zhang, S.~An, S.~Ma, and X.~Zhang, ``Constrained optimization with dynamic bound-scaling for effective {NLP} backdoor defense,'' in \emph{Proceedings of the 39th International Conference on Machine Learning (ICML)}, K.~Chaudhuri, S.~Jegelka, L.~Song, C.~Szepesvari, G.~Niu, and S.~Sabato, Eds., vol. 162.\hskip 1em plus 0.5em minus 0.4em\relax PMLR, 17--23 Jul 2022, pp. 19\,879--19\,892.

\bibitem{Wallace2019Triggers}
E.~Wallace, S.~Feng, N.~Kandpal, M.~Gardner, and S.~Singh, ``Universal adversarial triggers for attacking and analyzing {NLP},'' in \emph{Empirical Methods in Natural Language Processing (EMNLP)}, 2019.

\bibitem{fang2024data}
A.~Fang, A.~M. Jose, A.~Jain, L.~Schmidt, A.~T. Toshev, and V.~Shankar, ``Data filtering networks,'' in \emph{The Twelfth International Conference on Learning Representations (ICLR)}, 2024.

\bibitem{xu2024demystifying}
H.~Xu, S.~Xie, X.~Tan, P.-Y. Huang, R.~Howes, V.~Sharma, S.-W. Li, G.~Ghosh, L.~Zettlemoyer, and C.~Feichtenhofer, ``Demystifying {CLIP} data,'' in \emph{The Twelfth International Conference on Learning Representations (ICLR)}, 2024.

\bibitem{gadre2023datacomp}
S.~Y. Gadre, G.~Ilharco, A.~Fang, J.~Hayase, G.~Smyrnis, T.~Nguyen, R.~Marten, M.~Wortsman, D.~Ghosh, J.~Zhang, E.~Orgad, R.~Entezari, G.~Daras, S.~M. Pratt, V.~Ramanujan, Y.~Bitton, K.~Marathe, S.~Mussmann, R.~Vencu, M.~Cherti, R.~Krishna, P.~W. Koh, O.~Saukh, A.~Ratner, S.~Song, H.~Hajishirzi, A.~Farhadi, R.~Beaumont, S.~Oh, A.~Dimakis, J.~Jitsev, Y.~Carmon, V.~Shankar, and L.~Schmidt, ``Datacomp: In search of the next generation of multimodal datasets,'' in \emph{Thirty-seventh Conference on Neural Information Processing Systems Datasets and Benchmarks Track (NeurIPS)}, 2023.

\bibitem{li2022supervision}
Y.~Li, F.~Liang, L.~Zhao, Y.~Cui, W.~Ouyang, J.~Shao, F.~Yu, and J.~Yan, ``Supervision exists everywhere: A data efficient contrastive language-image pre-training paradigm,'' in \emph{International Conference on Learning Representations (ICLR)}, 2022.

\bibitem{tschannen2025siglip}
M.~Tschannen, A.~Gritsenko, X.~Wang, M.~F. Naeem, I.~Alabdulmohsin, N.~Parthasarathy, T.~Evans, L.~Beyer, Y.~Xia, B.~Mustafa, O.~H\'enaff, J.~Harmsen, A.~Steiner, and X.~Zhai, ``Siglip 2: Multilingual vision-language encoders with improved semantic understanding, localization, and dense features,'' \emph{arXiv preprint arXiv:2502.14786}, 2025.

\bibitem{Zhang2024LongCLIPUT}
B.~Zhang, P.~Zhang, X.~wen Dong, Y.~Zang, and J.~Wang, ``Long-clip: Unlocking the long-text capability of clip,'' in \emph{European Conference on Computer Vision (ECCV)}, 2024.

\bibitem{li2023blip2}
J.~Li, D.~Li, S.~Savarese, and S.~Hoi, ``Blip-2: bootstrapping language-image pre-training with frozen image encoders and large language models,'' in \emph{Proceedings of the 40th International Conference on Machine Learning (ICML)}, 2023.

\bibitem{dai2023instructblip}
W.~Dai, J.~Li, D.~Li, A.~Tiong, J.~Zhao, W.~Wang, B.~Li, P.~Fung, and S.~Hoi, ``Instruct{BLIP}: Towards general-purpose vision-language models with instruction tuning,'' in \emph{Thirty-seventh Conference on Neural Information Processing Systems (NeurIPS)}, 2023.

\bibitem{liu2023visual_llava}
H.~Liu, C.~Li, Q.~Wu, and Y.~J. Lee, ``Visual instruction tuning,'' in \emph{Proceedings of the 37th International Conference on Neural Information Processing Systems (NeurIPS)}.\hskip 1em plus 0.5em minus 0.4em\relax Red Hook, NY, USA: Curran Associates Inc., 2023.

\bibitem{li2023otter}
B.~Li, Y.~Zhang, L.~Chen, J.~Wang, F.~Pu, J.~A. Cahyono, J.~Yang, C.~Li, and Z.~Liu, ``Otter: A multi-modal model with in-context instruction tuning,'' \emph{IEEE Transactions on Pattern Analysis and Machine Intelligence (TPAMI)}, vol.~47, no.~9, pp. 7543--7557, 2025.

\bibitem{internlmxcomposer}
P.~Zhang, X.~Dong, B.~Wang, Y.~Cao, C.~Xu, L.~Ouyang, Z.~Zhao, S.~Ding, S.~Zhang, H.~Duan, W.~Zhang, H.~Yan, X.~Zhang, W.~Li, J.~Li, K.~Chen, C.~He, X.~Zhang, Y.~Qiao, D.~Lin, and J.~Wang, ``Internlm-xcomposer: A vision-language large model for advanced text-image comprehension and composition,'' \emph{arXiv preprint arXiv:2309.15112}, 2023.

\bibitem{Qwen-VL}
J.~Bai, S.~Bai, S.~Yang, S.~Wang, S.~Tan, P.~Wang, J.~Lin, C.~Zhou, and J.~Zhou, ``Qwen-vl: A versatile vision-language model for understanding, localization, text reading, and beyond,'' \emph{arXiv preprint arXiv:2308.12966}, 2023.

\bibitem{gu2017identifying}
T.~Gu, B.~Dolan-Gavitt, and S.~Garg, ``Identifying vulnerabilities in the machine learning model supply chain,'' in \emph{Proceedings of the Neural Information Processing Symposium Workshop Mach. Learning Security (MLSec)}, 2017, pp. 1--5.

\bibitem{chen2017targeted}
X.~Chen, C.~Liu, B.~Li, K.~Lu, and D.~Song, ``Targeted backdoor attacks on deep learning systems using data poisoning,'' \emph{arXiv preprint arXiv:1712.05526}, 2017.

\bibitem{liu2020reflection}
Y.~Liu, X.~Ma, J.~Bailey, and F.~Lu, ``Reflection backdoor: A natural backdoor attack on deep neural networks,'' in \emph{European Conference on Computer Vision (ECCV)}.\hskip 1em plus 0.5em minus 0.4em\relax Springer, 2020, pp. 182--199.

\bibitem{nguyen2021wanet}
T.~A. Nguyen and A.~T. Tran, ``Wanet - imperceptible warping-based backdoor attack,'' in \emph{International Conference on Learning Representations (ICLR)}, 2021.

\bibitem{li2021invisible}
Y.~Li, Y.~Li, B.~Wu, L.~Li, R.~He, and S.~Lyu, ``Invisible backdoor attack with sample-specific triggers,'' in \emph{Proceedings of the IEEE/CVF international conference on computer vision (ICCV)}, 2021, pp. 16\,463--16\,472.

\bibitem{Turner2019LabelConsistentBA}
A.~Turner, D.~Tsipras, and A.~Madry, ``Label-consistent backdoor attacks,'' \emph{arXiv preprint arXiv:1912.02771}, 2019.

\bibitem{Dai2019ABA}
J.~Dai, C.~Chen, and Y.~Li, ``A backdoor attack against lstm-based text classification systems,'' \emph{IEEE Access}, vol.~7, pp. 138\,872--138\,878, 2019.

\bibitem{kurita-etal-2020-weight}
K.~Kurita, P.~Michel, and G.~Neubig, ``Weight poisoning attacks on pretrained models,'' in \emph{Proceedings of the 58th Annual Meeting of the Association for Computational Linguistics (ACL)}, Online, Jul. 2020, pp. 2793--2806.

\bibitem{Chen2020BadNLBA}
X.~Chen, A.~Salem, D.~Chen, M.~Backes, S.~Ma, Q.~Shen, Z.~Wu, and Y.~Zhang, ``Badnl: Backdoor attacks against nlp models with semantic-preserving improvements,'' \emph{Proceedings of the 37th Annual Computer Security Applications Conference (ACSAC)}, 2020.

\bibitem{Qi2021HiddenKI}
F.~Qi, M.~Li, Y.~Chen, Z.~Zhang, Z.~Liu, Y.~Wang, and M.~Sun, ``Hidden killer: Invisible textual backdoor attacks with syntactic trigger,'' in \emph{Annual Meeting of the Association for Computational Linguistics (ACL)}, 2021.

\bibitem{Jia2021BadEncoderBA}
J.~Jia, Y.~Liu, and N.~Z. Gong, ``Badencoder: Backdoor attacks to pre-trained encoders in self-supervised learning,'' \emph{2022 IEEE Symposium on Security and Privacy (SP)}, pp. 2043--2059, 2021.

\bibitem{WANG2024103855}
Q.~Wang, C.~Yin, L.~Fang, Z.~Liu, R.~Wang, and C.~Lin, ``Ghostencoder: Stealthy backdoor attacks with dynamic triggers to pre-trained encoders in self-supervised learning,'' \emph{Computers \& Security}, vol. 142, p. 103855, 2024.

\bibitem{10646825}
G.~Tao, Z.~Wang, S.~Feng, G.~Shen, S.~Ma, and X.~Zhang, ``Distribution preserving backdoor attack in self-supervised learning,'' in \emph{2024 IEEE Symposium on Security and Privacy (SP)}, 2024, pp. 2029--2047.

\bibitem{Liang_2024_CVPR}
S.~Liang, M.~Zhu, A.~Liu, B.~Wu, X.~Cao, and E.-C. Chang, ``Badclip: Dual-embedding guided backdoor attack on multimodal contrastive learning,'' in \emph{Proceedings of the IEEE/CVF Conference on Computer Vision and Pattern Recognition (CVPR)}, June 2024, pp. 24\,645--24\,654.

\bibitem{Chen2019DeepInspectAB}
H.~Chen, C.~Fu, J.~Zhao, and F.~Koushanfar, ``Deepinspect: A black-box trojan detection and mitigation framework for deep neural networks,'' in \emph{International Joint Conference on Artificial Intelligence (IJCAI)}, 2019.

\bibitem{Chen2018DetectingBA}
B.~Chen, W.~Carvalho, N.~Baracaldo, H.~Ludwig, B.~Edwards, T.~Lee, I.~Molloy, and B.~Srivastava, ``Detecting backdoor attacks on deep neural networks by activation clustering,'' \emph{arXiv preprint arXiv:1811.03728}, 2018.

\bibitem{Tao2022BetterTI}
G.~Tao, G.~Shen, Y.~Liu, S.~An, Q.~Xu, S.~Ma, and X.~Zhang, ``Better trigger inversion optimization in backdoor scanning,'' \emph{2022 IEEE/CVF Conference on Computer Vision and Pattern Recognition (CVPR)}, pp. 13\,358--13\,368, 2022.

\bibitem{10.5555/3540261.3541554}
D.~Wu and Y.~Wang, ``Adversarial neuron pruning purifies backdoored deep models,'' in \emph{Proceedings of the 35th International Conference on Neural Information Processing Systems (NeurIPS)}.\hskip 1em plus 0.5em minus 0.4em\relax Red Hook, NY, USA: Curran Associates Inc., 2021.

\bibitem{Xu2019DetectingAT}
X.~Xu, Q.~Wang, H.~Li, N.~Borisov, C.~A. Gunter, and B.~Li, ``Detecting ai trojans using meta neural analysis,'' \emph{2021 IEEE Symposium on Security and Privacy (SP)}, pp. 103--120, 2019.

\bibitem{Wang2019NeuralCI}
B.~Wang, Y.~Yao, S.~Shan, H.~Li, B.~Viswanath, H.~Zheng, and B.~Y. Zhao, ``Neural cleanse: Identifying and mitigating backdoor attacks in neural networks,'' \emph{2019 IEEE Symposium on Security and Privacy (SP)}, pp. 707--723, 2019.

\bibitem{10.1145/3319535.3363216}
Y.~Liu, W.-C. Lee, G.~Tao, S.~Ma, Y.~Aafer, and X.~Zhang, ``Abs: Scanning neural networks for back-doors by artificial brain stimulation,'' in \emph{Proceedings of the 2019 ACM SIGSAC Conference on Computer and Communications Security (CCS)}.\hskip 1em plus 0.5em minus 0.4em\relax New York, NY, USA: Association for Computing Machinery, 2019, p. 1265–1282.

\bibitem{ishmam2024semanticshielddefendingvisionlanguage}
A.~M. Ishmam and C.~Thomas, ``Semantic shield: Defending vision-language models against backdooring and poisoning via fine-grained knowledge alignment,'' 2024.

\bibitem{singh2024perturb}
N.~D. Singh, F.~Croce, and M.~Hein, ``Perturb and recover: Fine-tuning for effective backdoor removal from clip,'' \emph{arXiv preprint arXiv:2412.00727}, 2024.

\bibitem{Liang2024UnlearningBT}
S.~Liang, K.~Liu, J.~Gong, J.~Liang, Y.~Xun, E.-C. Chang, and X.~Cao, ``Unlearning backdoor threats: Enhancing backdoor defense in multimodal contrastive learning via local token unlearning,'' \emph{ArXiv}, 2024.

\bibitem{Bansal2023CleanCLIPMD}
H.~Bansal, N.~Singhi, Y.~Yang, F.~Yin, A.~Grover, and K.-W. Chang, ``Cleanclip: Mitigating data poisoning attacks in multimodal contrastive learning,'' \emph{2023 IEEE/CVF International Conference on Computer Vision (ICCV)}, pp. 112--123, 2023.

\bibitem{11094745}
J.~Kim, E.~Esmaeili, and Q.~Qiu, ``Text embedding is not all you need: Attention control for text-to-image semantic alignment with text self-attention maps,'' in \emph{2025 IEEE/CVF Conference on Computer Vision and Pattern Recognition (CVPR)}, 2025, pp. 8031--8040.

\bibitem{xiao2024efficient}
G.~Xiao, Y.~Tian, B.~Chen, S.~Han, and M.~Lewis, ``Efficient streaming language models with attention sinks,'' in \emph{The Twelfth International Conference on Learning Representations (ICLR)}, 2024.

\bibitem{Wang2022DiffusionDBAL}
Z.~J. Wang, E.~Montoya, D.~Munechika, H.~Yang, B.~Hoover, and D.~H. Chau, ``{D}iffusion{DB}: A large-scale prompt gallery dataset for text-to-image generative models,'' in \emph{Proceedings of the 61st Annual Meeting of the Association for Computational Linguistics (ACL)}.\hskip 1em plus 0.5em minus 0.4em\relax Toronto, Canada: Association for Computational Linguistics, Jul. 2023, pp. 893--911.

\bibitem{OpenAI}
``Openai. hello gpt-4o,'' \url{https://openai.com/index/hello-gpt-4o/}.

\bibitem{gal2023an}
R.~Gal, Y.~Alaluf, Y.~Atzmon, O.~Patashnik, A.~H. Bermano, G.~Chechik, and D.~Cohen-or, ``An image is worth one word: Personalizing text-to-image generation using textual inversion,'' in \emph{The Eleventh International Conference on Learning Representations (ICLR)}, 2023.

\bibitem{UAP}
S.-M. Moosavi-Dezfooli, A.~Fawzi, O.~Fawzi, and P.~Frossard, ``Universal adversarial perturbations,'' in \emph{2017 IEEE Conference on Computer Vision and Pattern Recognition (CVPR)}, 2017, pp. 86--94.

\bibitem{10.5555/3327345.3327535}
H.~Li, Z.~Xu, G.~Taylor, C.~Studer, and T.~Goldstein, ``Visualizing the loss landscape of neural nets,'' in \emph{Proceedings of the 32nd International Conference on Neural Information Processing Systems (NeurIPS)}.\hskip 1em plus 0.5em minus 0.4em\relax Red Hook, NY, USA: Curran Associates Inc., 2018, p. 6391–6401.

\bibitem{10.5555/3600270.3600632}
G.~Cui, L.~Yuan, B.~He, Y.~Chen, Z.~Liu, and M.~Sun, ``A unified evaluation of textual backdoor learning: frameworks and benchmarks,'' in \emph{Proceedings of the 36th International Conference on Neural Information Processing Systems (NeurIPS)}.\hskip 1em plus 0.5em minus 0.4em\relax Red Hook, NY, USA: Curran Associates Inc., 2022.

\bibitem{Addsent}
J.~Dai, C.~Chen, and Y.~Li, ``A backdoor attack against lstm-based text classification systems,'' \emph{IEEE Access}, vol.~7, pp. 138\,872--138\,878, 2019.

\bibitem{Lin2014MicrosoftCC}
T.-Y. Lin, M.~Maire, S.~J. Belongie, J.~Hays, P.~Perona, D.~Ramanan, P.~Doll{\'a}r, and C.~L. Zitnick, ``Microsoft coco: Common objects in context,'' in \emph{Proceedings of the European Conference on Computer Vision (ECCV)}, 2014.

\end{thebibliography}

\begin{IEEEbiography}[{\includegraphics[width=1in,height=1.25in,clip,keepaspectratio]{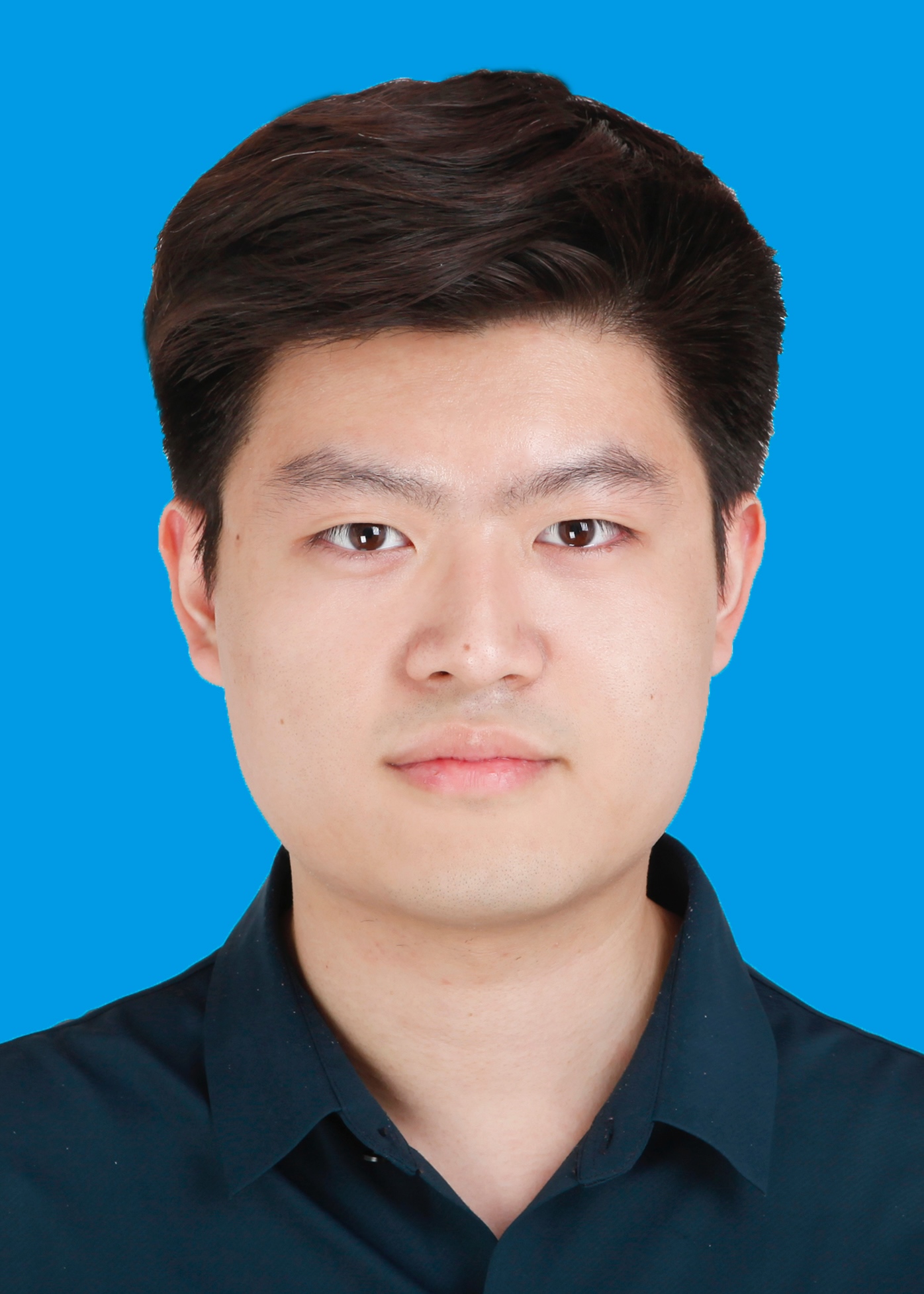}}]{Zhongqi Wang} (Student Member, IEEE) received the BS degree in artificial intelligence from Beijing Institute of Technology, in 2023. He is currently working toward the Ph.D. degree with the Institute of Computing Technology (ICT), Chinese Academy of Sciences (CAS). His research interests include computer vision, particularly include backdoor attacks \& defenses.
\end{IEEEbiography}

\begin{IEEEbiography}[{\includegraphics[width=1in,height=1.25in,clip,keepaspectratio]{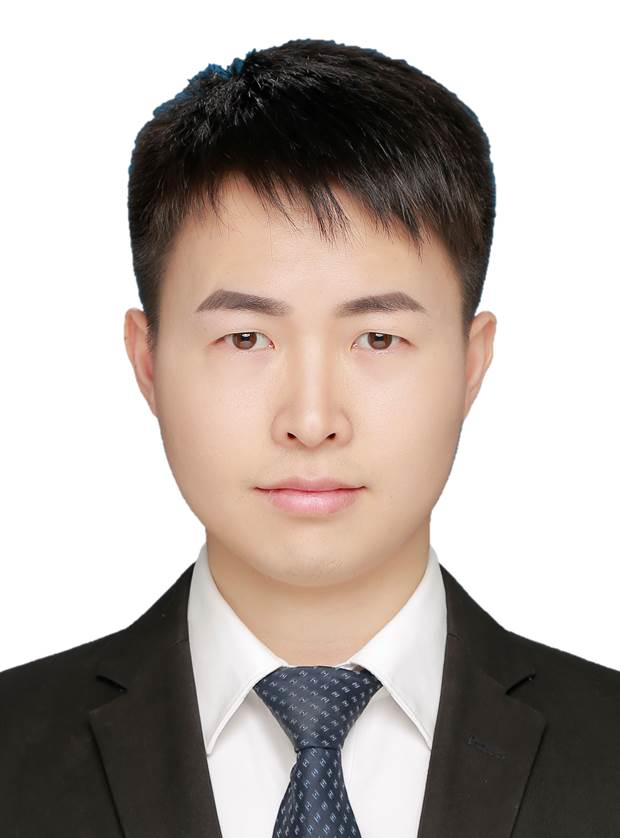}}]{Jie Zhang}
(Member, IEEE) received the Ph.D. degree from the University of Chinese Academy of Sciences (CAS), Beijing, China. He is currently an Associate Professor with the Institute of Computing Technology, CAS. His research interests include computer vision, pattern recognition, machine learning, particularly include adversarial attacks and defenses,  domain generalization, AI safety and trustworthiness.
\end{IEEEbiography}

\begin{IEEEbiography}[{\includegraphics[width=1in,height=1.25in,clip,keepaspectratio]{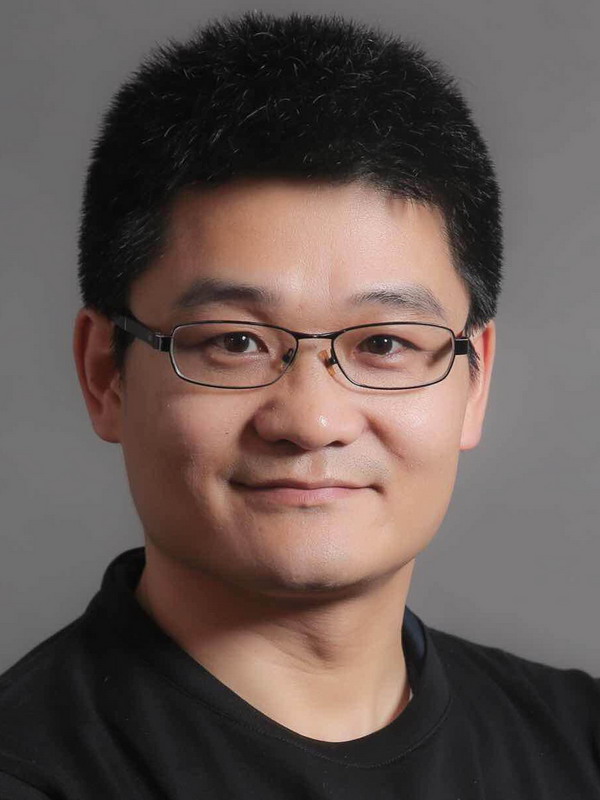}}]{Shiguang Shan}
(Fellow, IEEE) received the Ph.D. degree in computer science from the Institute of Computing Technology (ICT), Chinese Academy of Sciences (CAS), Beijing, China, in 2004. He has been a Full Professor with ICT since 2010, where he is currently the Director of the Key Laboratory of Intelligent Information Processing, CAS. His research interests include signal processing, computer vision, pattern recognition, and machine learning. He has published more than 300 articles in related areas. He served as the General Co-Chair for IEEE Face and Gesture Recognition 2023, the General Co-Chair for Asian Conference on Computer Vision (ACCV) 2022, and the Area Chair of many international conferences, including CVPR, ICCV, AAAI, IJCAI, ACCV, ICPR, and FG. He was/is an Associate Editors of several journals, including IEEE Transactions on Image Processing, Neurocomputing, CVIU, and PRL. He was a recipient of the China's State Natural Science Award in 2015 and the China’s State S\&T Progress Award in 2005 for his research work.
\end{IEEEbiography}

\begin{IEEEbiography}[{\includegraphics[width=1in,height=1.25in,clip,keepaspectratio]{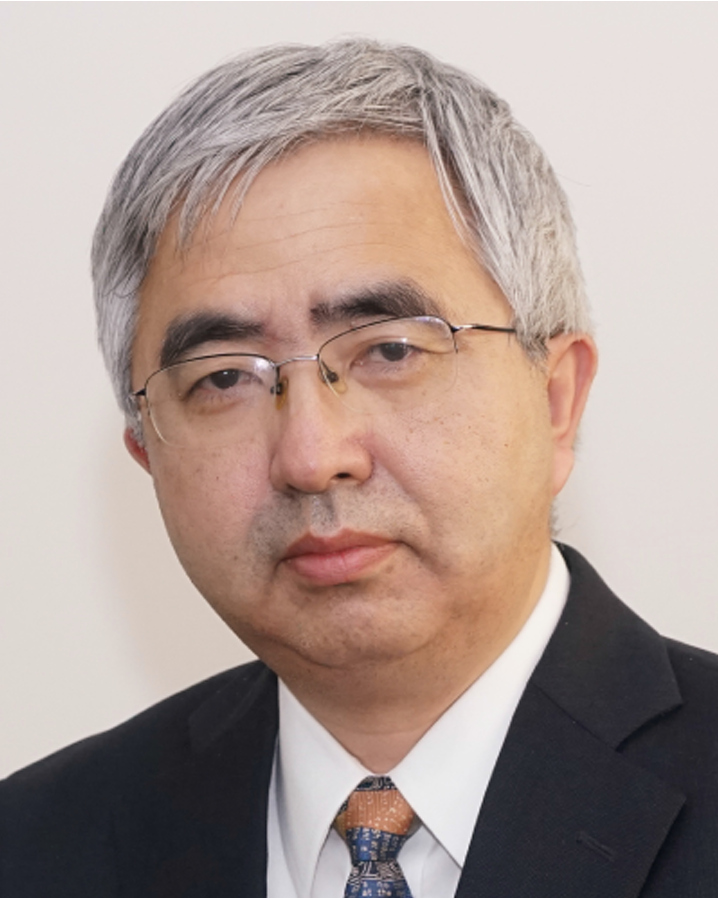}}]{Xilin Chen} (Fellow, IEEE) is currently a Professor with the Institute of Computing Technology, Chinese
 Academy of Sciences (CAS). He has authored one
 book and more than 400 articles in refereed journals
 and proceedings in the areas of computer vision,
 pattern recognition, image processing, and multi
modal interfaces. He is a fellow of the ACM,
 IAPR, and CCF. He is also an Information Sciences
 Editorial Board Member of Fundamental Research,
 an Editorial Board Member of Research, a Senior
 Editor of the Journal of Visual Communication and
 Image Representation, and an Associate Editor-in-Chief of the Chinese Jour
nal of Computers and Chinese Journal of Pattern Recognition and Artificial
 Intelligence. He served as an organizing committee member for multiple
 conferences, including the General Co-Chair of FG 2013/FG 2018, VCIP
 2022, the Program Co-Chair of ICMI 2010/FG 2024, and an Area Chair of
 ICCV/CVPR/ECCV/NeurIPS for more than ten times.
\end{IEEEbiography}

\clearpage
\appendices
\section*{Supplementary Material}
We provide the following supplementary materials in the Appendix, including the additional details and analysis on our method.

\subsection{Reproducibility} \label{Reproducibility}

 \name{} is executed on Ubuntu 20.04.3 LTS with an Intel(R) Xeon(R) Platinum 8358P CPU @ 2.60GHz. The machine is equipped with 1.0 TB of RAM and 8 Nvidia RTX4090-24GB GPUs. Our experiments are conducted using CUDA 12.2, Python 3.10.0, and PyTorch 2.2.0. 

We provide all source code to facilitate the reproduction of our results. The code is available at \url{https://github.com/Robin-WZQ/AMDET}. All configuration files and training and evaluation scripts for \name{} are included in the repository. 

\subsection{The structure of the loss landscape} \label{sec:loss landscape}

We denote the loss function as $\mathcal{L}(v)$, where $v$ represents the implicit backdoor feature to be optimized. Let $v^*$ denote the optimized feature. Here, we analyze the variation of the loss in a two-dimensional perturbation subspace:  
\begin{equation}
\mathcal{H}(v^*) = \mathcal{L}(v^* + \Delta v),
\end{equation}
where $\Delta v = \alpha \cdot \delta + \beta \cdot \eta$ is a perturbation spanned by two orthogonal directions $\delta, \eta$.  

\underline{\textbf{Case 1:}} For backdoor models, the feature $v$ is explicitly optimized during training to align with the target representation. Thus, the obtained feature $v^*$ can be regarded as a local optimum of the training objective.  

Expanding $\mathcal{L}$ in a second-order Taylor series around $v^*$:  
\begin{equation}
\mathcal{L}(v^* + \Delta v) \approx \mathcal{L}(v^*) + \nabla \mathcal{L}(v^*)^\top \Delta v + \frac{1}{2} \Delta v^\top H(v^*) \Delta v, 
\label{taylor}
\end{equation}
where $H(v^*)$ is the Hessian of $\mathcal{L}$ at $v^*$, \textit{i.e.}, a $d \times d$ symmetric matrix.  

If $v^*$ is indeed a local minimizer, then $\nabla \mathcal{L}(v^*) \approx 0$. $H(v^*)$ is positive definite, \textit{i.e.}, $\Delta v^\top H(v^*) \Delta v > 0$ for all $\Delta v \neq 0$.  

Therefore, Eq.~(\ref{taylor}) reduces to:  
\begin{equation}
\mathcal{L}(v^* + \Delta v) \approx \mathcal{L}(v^*) + \frac{1}{2} \Delta v^\top H(v^*) \Delta v.  
\end{equation}

Here, $\mathcal{L}(v^*)$ is a constant, and the quadratic form $\Delta v^\top H(v^*) \Delta v$ yields a smooth ellipsoidal bowl in the subspace. This explains the smooth and symmetric landscapes observed in experiments for backdoor models.

\underline{\textbf{Case 2:}} For benign models, the feature $v^*$ is obtained post-hoc during inversion, rather than optimized explicitly during training. In this case, $v^*$ is not necessarily a local minimizer of $\mathcal{L}$, thus $\nabla \mathcal{L}(v^*) \not\approx 0$. Besides, the spectrum of $H(v^*)$ may exhibit unstable.  

As a result, the linear term $\nabla \mathcal{L}(v^*)^\top \Delta v$ introduces directional bias, and the quadratic term reflects irregular curvature. The loss landscape therefore becomes asymmetric and sensitive to perturbations.

\para{Remark.} This analysis explains the empirical difference observed in Fig.~\ref{fig:loss_landscape}: Benign models exhibit \emph{natural backdoor feature} with irregular landscapes, while backdoor models show bowl-shaped quadratic structures.

\subsection{Instruction Template} \label{sec:template}

\begin{figure}[t]
    \centering
    \includegraphics[width=0.9\linewidth]{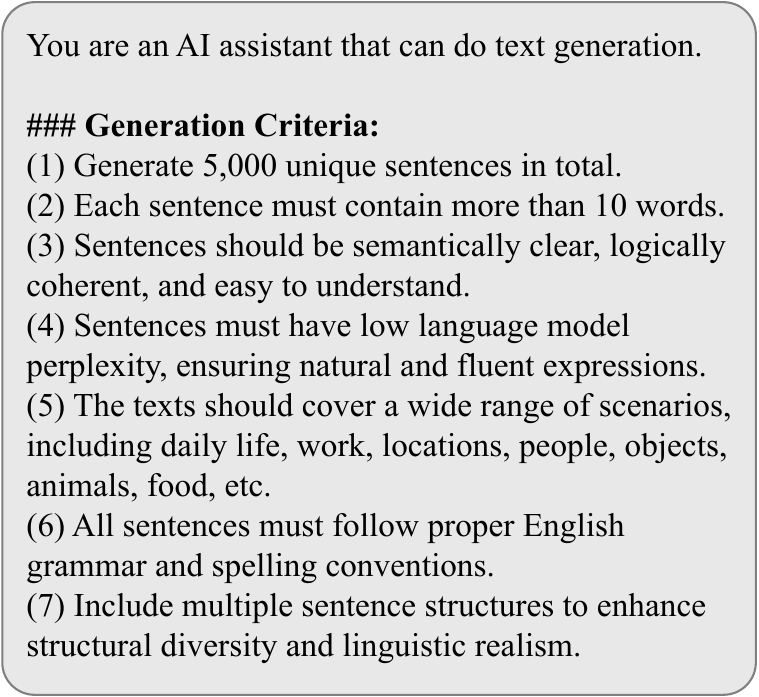}
    \caption{The instruction template for generative task.}
    \label{fig:instruction}
\end{figure}

Fig.~\ref{fig:instruction} illustrates the instruction template provided to ChatGPT for generating the shallow dataset $\mathcal{P}$ used in backdoor detection.
The goal is to encourage ChatGPT to generate linguistically diverse samples.

\subsection{Proof of Proposition 1} \label{sec:proof}

\para{Proposition 1.}  
\textit{Define a matrix $R$ as}
\begin{equation}
R_{i,j} = {\mathbf{e}_i^{(l)}}^{\top} {W_v^{(l,h)}}^{\top} W_v^{(l,h)} \mathbf{e}_j^{(l)},
\end{equation}
\textit{where $\mathbf{e}_i^{(l)}$ denotes the $i$-th token embedding at layer $l$, 
and $W_v^{(l,h)}$ is the value projection matrix in the $h$-th attention head. 
Let $t$ denote the index of the attention concentration token. 
}
\textit{Suppose for benign samples it has the property}
\begin{equation}
\frac{|R_{mn}|}{R_{tt}} \sim \mathcal{O}\!\left(\frac{1}{\epsilon}\right), 
\frac{|R_{tm}|}{R_{tt}} \sim \mathcal{O}(1),
\ m \neq t, n \neq t,
\end{equation}
\textit{where $O(\epsilon)$ mean terms that are linear or higher order in $\epsilon$. For backdoor samples it has}
\begin{equation}
\frac{|R_{mn}|}{R_{tt}} \sim \mathcal{O}(1), 
\frac{|R_{tm}|}{R_{tt}} \sim \mathcal{O}(1), 
\ m \neq t, n \neq t,
\end{equation}
\textit{and}
\begin{align}
\epsilon = \frac{\sum_{j \neq t} \bar{\mathcal{M}}_{ij}}{\bar{\mathcal{M}}_{it}} \ll 1, \quad i \neq t,\\
\epsilon_{\text{backdoor}} < \epsilon_{\text{benign}}.
\end{align}

\textit{Then, the following holds:}
\begin{equation}
Sim_{X}^{Backdoor}>Sim_{X}^{Benign}.
\end{equation}

\textit{proof. } Let $\mathcal{M}_{ij}$ be the attention weight from token $i$ to $j$, with $t$ denoting the concentration token. The self-attention output for token $i$ is
\begin{equation}
\mathbf{o}_i^{(\ell,h)} = \sum_{j=1}^{s} \mathcal{M}_{ij}^{(\ell,h)} W_v^{(\ell,h)} \mathbf{e}_j^{(\ell)}.
\end{equation}

\para{Lemma 1.}(Cosine similarity preservation \cite{11094745})
\textit{Let the full self-attention layer include an output linear projection $W_{\text{out}}$ and residual:}
\begin{align}
   \mathbf{e}_i^{(\ell)\text{out}} &= \mathbf{e}_i^{(\ell)} + \mathbf{e}_i^{\prime(\ell)},\\
\mathbf{e}_i^{\prime(\ell)} &= \text{SelfAttention}_i(\mathbf{e}_1^{(\ell)},\dots,\mathbf{e}_s^{(\ell)}) W_{\text{out}}. 
\end{align}

\textit{If $\mathbf{e}_i^{\prime(\ell)} = \mathcal{O}(\epsilon)$ relative to $\mathbf{e}_i^{(\ell)}$, then}
\begin{equation}
\cos(\mathbf{e}_i^{(\ell)\text{out}}, \mathbf{e}_j^{(\ell)\text{out}}) 
= \cos(\mathbf{e}_i^{(\ell)} + \mathbf{e}_i^{\prime(\ell)}, \mathbf{e}_j^{(\ell)} + \mathbf{e}_j^{\prime(\ell)}) + \mathcal{O}(\epsilon^2),
\end{equation}
\textit{i.e., the output cosine is close to the input cosine up to second-order corrections.}

The pairwise output cosine similarity is
\begin{equation}
\cos(\mathbf{o}_i,\mathbf{o}_j) = \frac{\sum_{m,n} \mathcal{M}_{im}\mathcal{M}_{jn} R_{mn}}{\sqrt{\sum_{m,n} \mathcal{M}_{im}\mathcal{M}_{in} R_{mn}}\sqrt{\sum_{m,n} \mathcal{M}_{jm}\mathcal{M}_{jn} R_{mn}}}.
\end{equation}

We can rewrite the numerator as:
\begin{align}
\sum_{m,n} \mathcal{M}_{im}\mathcal{M}_{jn} R_{mn} &= \mathcal{M}_{it}\mathcal{M}_{jt} R_{tt} 
+ \sum_{m\neq t} \mathcal{M}_{im} \mathcal{M}_{jt} R_{mt} \\
&+ \sum_{n\neq t} \mathcal{M}_{it} \mathcal{M}_{jn} R_{tn} \\
&+ \sum_{m\neq t,n\neq t} \mathcal{M}_{im} \mathcal{M}_{jn} R_{mn}.
\end{align}

Using the attention scaling assumptions
\[
\mathcal{M}_{it} = 1 - \mathcal{O}(\epsilon), \quad \mathcal{M}_{im} = \mathcal{O}(\epsilon) \text{ for } m \neq t,
\]
For benign samples, we can obtain:
\begin{align}
\sum_{m,n} \mathcal{M}_{im}\mathcal{M}_{jn} R_{mn} &=  (1 - \mathcal{O}(\epsilon))^2 R_{tt}\\ 
&+ \sum_{m\neq t} \mathcal{O}(\epsilon) \cdot (1-\mathcal{O}(\epsilon)) \cdot \mathcal{O}(R_{tt})  \\
&+ \mathcal{O}(\epsilon R_{tt})
+ \sum_{m,n\neq t} \mathcal{O}(\epsilon^2) \cdot \mathcal{O}(R_{mn}).
\end{align}

Thus we get for benign samples:
\begin{equation}
\sum_{m,n} \mathcal{M}_{im} \mathcal{M}_{jn} R_{mn} = R_{tt} + \mathcal{O}(\epsilon R_{tt}).
\end{equation}

For backdoor samples, because $R_{mn} \sim \mathcal{O}(1)$ for $m,n\neq t$, the cross-terms with $m,n\neq t$ become
\[
\sum_{m\neq t,n\neq t} \mathcal{M}_{im} \mathcal{M}_{jn} R_{mn} = \sum \mathcal{O}(\epsilon^2) \cdot \mathcal{O}(1) = \mathcal{O}(\epsilon^2),
\]
while the other terms are still $\mathcal{O}(\epsilon^2)$ or smaller, giving:
\begin{equation}
\sum_{m,n} \mathcal{M}_{im} \mathcal{M}_{jn} R_{mn} = R_{tt} + \mathcal{O}(\epsilon^2 R_{tt}).
\end{equation}

Similarly, the denominator of the cosine similarity, \textit{i.e.},
\[
\|\mathbf{o}_i\| \|\mathbf{o}_j\| = \sqrt{\sum_{m,n} \mathcal{M}_{im} \mathcal{M}_{in} R_{mn}} \sqrt{\sum_{m,n} \mathcal{M}_{jm} \mathcal{M}_{jn} R_{mn}},
\]
has the same leading term $\mathcal{M}_{it}^2 R_{tt}$ with the same order corrections: $\mathcal{O}(\epsilon R_{tt})$ for benign and $\mathcal{O}(\epsilon^2 R_{tt})$ for backdoor.

We can then write the cosine similarity in a compact form by factoring out the leading term:
\begin{align}
\cos(\mathbf{o}_i,\mathbf{o}_j) = \frac{1 + \tilde\delta_{ij}}{\sqrt{(1+\tilde\eta_i)(1+\tilde\eta_j)}}, \quad \\
\tilde\delta_{ij},\tilde\eta_i,\tilde\eta_j = \mathcal{O}(\epsilon) \text{ (benign) or } \mathcal{O}(\epsilon^2) \text{ (backdoor)}.
\end{align}

Expanding the denominator using a Taylor expansion $\sqrt{1+x} \approx 1 + x/2$ and keeping terms up to the first non-zero order in $\epsilon$, we obtain:
\begin{equation}
\cos(\mathbf{o}_i,\mathbf{o}_j) \approx 1 + \tilde\delta_{ij} - \frac{1}{2}(\tilde\eta_i+\tilde\eta_j) + \mathcal{O}(\epsilon^2).
\end{equation}

Hence, we see that:
\begin{itemize}
    \item For benign samples, the leading-order correction is $\mathcal{O}(\epsilon)$.
    \item For backdoor samples, the leading-order correction is $\mathcal{O}(\epsilon^2)$.
\end{itemize}

Since $\epsilon_{\text{backdoor}} < \epsilon_{\text{benign}}$, it follows that
\begin{equation}
\cos(\mathbf{o}_i,\mathbf{o}_j)_{\text{backdoor}} > \cos(\mathbf{o}_i,\mathbf{o}_j)_{\text{benign}}.
\end{equation}

By Lemma 1, for the full layer including $W_{\text{out}}$ and residual:
\begin{equation}
\cos(\mathbf{e}_i^{(\ell)\text{out}},\mathbf{e}_j^{(\ell)\text{out}})_{\text{backdoor}}
> \cos(\mathbf{e}_i^{(\ell)\text{out}},\mathbf{e}_j^{(\ell)\text{out}})_{\text{benign}} + \mathcal{O}(\epsilon^2).
\end{equation}

Equivalently,

\begin{equation}
Sim_X^{\text{Backdoor}} > Sim_X^{\text{Benign}}.
\end{equation}

\qed

\vfill

\end{document}